\documentclass[conference]{IEEEtran}
\usepackage{times}

\usepackage[numbers]{natbib}
\usepackage{multicol}
\usepackage[bookmarks=true]{hyperref}
\usepackage{amsmath}
\usepackage{graphicx} 
\usepackage{amssymb}
\usepackage{multirow}
\usepackage{xspace}
\usepackage{comment}
\usepackage{algorithm}
\usepackage{algpseudocode}
\usepackage{subcaption}
\usepackage[table,xcdraw]{xcolor}
\usepackage{wrapfig}

\definecolor{scarletred}{rgb}{0.8, 0.0, 0.2}
\hypersetup{
  colorlinks = true,  
  urlcolor   = scarletred,  
  linkcolor  = blue,  
  citecolor  = red,  
}

\newcommand{\stela}{{\tt STELA}\xspace}
\newcommand{\steap}{{\tt STEAP}\xspace}
\newcommand{\scate}{{\tt SCATE}\xspace}
\newcommand{\openLoop}{{\tt OPEN-LOOP}\xspace}
\newcommand{\replanning}{{\tt SBMP-REPLANNING}\xspace} 
\newcommand{\sbmp}{{\tt SBMP}\xspace}
\newcommand{\fg}{{\tt FG}\xspace}
\newcommand{\sbmps}{{\tt SBMPs}\xspace}
\newcommand{\fgs}{{\tt FGs}\xspace}
\newcommand{\mushr}{MuSHR\xspace}

\begin{document}

\title{\huge Kinodynamic Trajectory Following with \stela:\\ Simultaneous Trajectory Estimation \& Local Adaptation}

\author{
\authorblockN{Edgar Granados,
Sumanth Tangirala,
Kostas E. Bekris}
\authorblockA{Dept. of Computer Science, Rutgers University, NJ, USA\\ \{eg585, st1122, kb572\}@cs.rutgers.edu}
}


\twocolumn[{
\begin{@twocolumnfalse}

\maketitle   
    \centering
    \vspace{-0.2in}
    \captionsetup{type=figure}
	\includegraphics[width=0.95\textwidth]{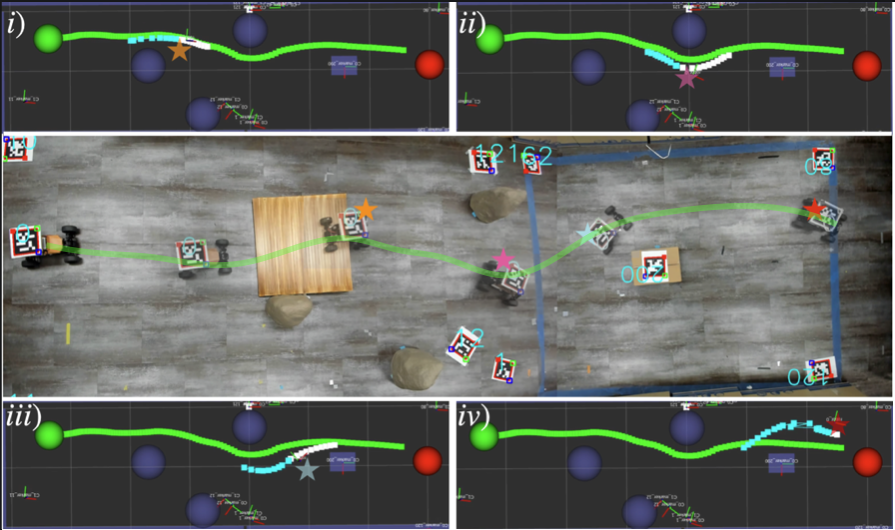}
    \captionof{figure}{\small Middle: \stela execution on a \emph{real} \mushr robot. The middle image is a composite from 2 top-down cameras used for localization, covering a 7.6mx2.3m workspace. The robot follows a trajectory computed by a planner with knowledge of the obstacles (rocks and boxes) but no knowledge of the ramp, affecting execution. Top and Bottom: i) \stela estimation and plan when the robot is on the unknown ramp; ii) the robot recovers from the ramp and avoids an obstacle; iii) \stela adapts the plan to follow the planned trajectory while avoiding another obstacle; iv) the robot reaches the end of the plan without collisions. Rviz visualization includes obstacles, planned trajectory (green), forward horizon (white), and history (cyan). Stars indicate corresponding states between the visualization and the real robot. }
    \label{fig:intro}
\end{@twocolumnfalse}
}]

\begin{abstract}
State estimation and control are often addressed separately, leading to unsafe execution due to sensing noise, execution errors, and discrepancies between the planning model and reality. Simultaneous control and trajectory estimation using probabilistic graphical models has been proposed as a unified solution to these challenges. Previous work, however, relies heavily on appropriate Gaussian priors and is limited to holonomic robots with linear time-varying models. The current research extends graphical optimization methods to vehicles with arbitrary dynamical models via Simultaneous Trajectory Estimation and Local Adaptation (\stela). The overall approach initializes feasible trajectories using a kinodynamic, sampling-based motion planner. Then, it simultaneously: (i) estimates the past trajectory based on noisy observations, and (ii) adapts the controls to be executed to minimize deviations from the planned, feasible trajectory, while avoiding collisions. The proposed factor graph representation of trajectories in \stela can be applied for any dynamical system given access to first or second-order state update equations, and introduces the duration of execution between two states in the trajectory discretization as an optimization variable. These features provide both generalization and flexibility in trajectory following. In addition to targeting computational efficiency, the proposed strategy performs incremental updates of the factor graph using the iSAM algorithm and introduces a time-window mechanism. This mechanism allows the factor graph to be dynamically updated to operate over a limited history and forward horizon of the planned trajectory. This enables online updates of controls at a minimum of 10Hz. Experiments demonstrate that \stela achieves at least comparable performance to previous frameworks on idealized vehicles with linear dynamics. \stela also directly applies to and successfully solves trajectory following problems for more complex dynamical models. Beyond generalization, simulations assess \stela's robustness under varying levels of sensing and execution noise, while ablation studies highlight the importance of different components of \stela. Real-world experiments validate \stela's practical applicability on a low-cost MuSHR robot, which exhibits high noise and non-trivial dynamics.\\
Website: \href{https://go.rutgers.edu/46618xjt}{https://go.rutgers.edu/46618xjt}
\end{abstract}
\IEEEpeerreviewmaketitle

\section{INTRODUCTION}

This paper focuses on achieving reliable simultaneous trajectory estimation and following for kinodynamic systems in static, partially modeled environments, under sensing and actuation uncertainty as well as reality gaps for the planning robot model.  It proposes Simultaneous Trajectory Estimation and Local Adaptation (\stela), a graphical optimization framework that builds on top of prior methods for simultaneous state estimation and control based on factor graph representations \cite{dellaert2017factor,king2022simultaneous}. It extends: (i) the efficacy of such approaches by increasing the success rate of returning feasible, collision-free trajectories, even under significant noise, (ii) their applicability to more general dynamical systems, and (iii) achieves improved computational efficiency.

{\bf Motivation:} Reliable mobile robot navigation, especially for low-cost platforms, such as the \mushr robot used in this work and shown in Fig.~\ref{fig:intro}, can be challenging due to observation noise, high actuation errors, and a significant reality gap of the underlying models.  For robots with significant dynamics, the planning models are often analytical dynamic expressions, which allow for fast computation but tend to be rough approximations. This results in significant deviations from the planned trajectory during execution. While system identification  \cite{johansson2000state,bahnemann2017sampling,sysIdGPTingfan} can reduce the model gap, it does not fully address it. This is often because the environment is also partially observable. For instance, a floor map indicating obstacles may be available, but not all aspects of the environment are modeled, such as friction coefficients and traversable features, such as ramps and speed obstructions. 

An approach to deal with the model gap is to use feedback controllers for trajectory following, given the latest state estimate \cite{hoffmann2007autonomous,paden2016survey}. However, observation and actuation noise can lead to errors in state estimation, where the focus is often filtering, i.e., estimating the latest robot pose incrementally. State estimation noise can compound to result in deviations in trajectory following. In addition, most trajectory following controllers ignore obstacles and are executed independently from the state estimation process, so significant trajectory deviations also lead to collisions. 

An alternative strategy is replanning online, e.g., with sampling-based motion planners (\sbmp) \cite{karaman2011sampling,littlefield2018efficient,kleinbort2020refined} or trajectory optimization planners \cite{ratliff2009chomp,toussaint2009robot,kalakrishnan2011stomp,toussaint2014newton}. \sbmps can eventually provide high-quality, feasible solutions, but typically it is not possible to achieve high-frequency replanning rates for dynamical systems (e.g., at or above 10Hz) to deal with the model gap without a trajectory follower. Optimization-based approaches can sometimes provide solutions fast, but due to their local nature, they may get stuck in local minima if not properly initialized and can be parameter sensitive.

{\bf Factor Graph Optimization:} A promising direction for addressing the above issues, which this paper builds on, is probabilistic graphical models based on factor graphs and corresponding optimization methods \cite{dellaert2017factor}. They can simultaneously solve trajectory estimation and control or planning challenges as a unified problem \cite{king2022simultaneous,mukadam2019steap}. These solutions perform smoothing instead of filtering, i.e., they estimate the entire most likely trajectory the robot has followed given all the available observations. Smoothing often leads to improved estimates relative to filtering. Smoothing solutions were traditionally slower to compute, but the progress with factor graph optimization tools has allowed such problems to be solved online. Furthermore, these methods are able to adapt the robot controls simultaneously to achieve collision avoidance by taking obstacles into account during the control optimization process, or alternatively, come up with new planning solutions. 

Nevertheless, most of the existing solutions in this space: (i) rely heavily on Gaussian priors regarding the underlying probabilistic processes, which may not reflect the true uncertainty of the system, and (ii) are limited to holonomic robots given linear-time varying models, reducing their applicability on non-holonomic vehicles with significant dynamics. Furthermore, it is also desirable to operate such solutions at high frequencies to maximize robustness to disturbances.

{\bf Proposed Method and Contribution:} The proposed \stela framework first calls an asymptotically optimal \sbmp for kinodynamic systems \cite{kleinbort2020refined, littlefield2018efficient} in order to acquire a feasible, collision-free trajectory given the available planning model. A key aspect of \stela is that it aligns the output of the \sbmp with the consecutive trajectory optimization via a common graphical representation. Factor Graphs (\fgs) are a natural interface for this purpose. In particular, the underlying motion graph produced by an \sbmp is transformed by the proposed approach into a \fg. Then, \stela uses the extracted \fg to simultaneously perform: (a) smoothing of the past trajectory given the latest observations, and (b) dynamically adapting the controls to be executed so as to minimize deviations from the \sbmp trajectory, while avoiding collisions.

The proposed \fg representation in \stela, illustrated in Fig.~\ref{fig:dynamics_fg}, is general in nature and can be applied to any dynamical system given access to first or second-order state update equations. It does not make any assumptions in terms of Gaussian priors for the underlying processes, and it does not require linear-time varying models, which limit applicability to idealistic holonomic vehicles. Instead, it only uses the solution achieved by the \sbmp as a prior and employs non-linear factors representing the system's dynamics. The proposed \fg also includes the duration of execution between two states in the trajectory discretization as a variable to be optimized.  This allows the optimizer to \textit{stretch} or \textit{contract} edges depending on the estimated state of the system. The combination of these features provides \emph{generalization}, in terms of the range of the dynamical systems that can be modeled, as well as \emph{high success rate} in finding a collision-free path even when the robot has deviated from the planned solution.

For increased \emph{computational efficiency}, the proposed strategy allows for the use of incremental updates of the \fg by: (a) using the iSAM2 optimizer, and also (b) introducing a sliding window mechanism. In particular, given the adopted \fg representation and seeking the Maximum a posteriori (MAP) solution via incremental inference, the iSAM2 optimizer incorporates high-frequency observations, adapting the underlying graph in a computationally efficient manner, obtaining the most likely estimate of the robot's past trajectory. The sliding window mechanism allows the factor graph to be dynamically updated at high frequency by operating over a limited past history and forward horizon of the planned trajectory.  The combination of these features enables online control updates to be generated at a minimum of 10Hz and on average at 20- 30Hz, depending on the setup.

{\bf Experimentation:} The framework is tested first in simulation for different environments and different levels of observation and actuation noise both for an idealized holonomic model used in prior work as well a second-order dynamical system not addressable by prior \fg work corresponding to a \mushr robot. The simulations demonstrate that \stela achieves at least comparable, and often superior, performance to previous frameworks on the idealized vehicle with linear dynamics. More critically, \stela also achieves a high success rate for the second-order dynamical system and good robustness given different noise levels. Ablation studies highlight the importance of different components of the approach, such as the importance of the \sbmp initialization, the introduction of control duration as an optimization variable, and the incorporation of the sliding window approach. Real-world experiments validate \stela's practical applicability on a low-cost \mushr robot that exhibits high noise and non-trivial dynamics.

\section{RELATED WORK} \label{sec:related_work}

{\bf Motion planning} consists of finding a plan for a robot to move in an environment from a starting state to a desired goal region without collisions. Multiple classifications of motion planning algorithms have been proposed \cite{Hauser2020,ortiz2024idb,atreya2022state,mukadam2018continuous}. This paper primarily focuses on Sampling-Based Motion Planners (\sbmp) and trajectory optimization. 

\sbmps build graphical representations of the underlying robot's state space and can provide guarantees, namely probabilistic completeness (PC) and asymptotic optimality (AO), for kinematic \cite{karaman2011sampling,gammell2015batch} and kinodynamic \cite{paden2020generalized,littlefield2018efficient,kleinbort2020refined} systems. Additionally, \sbmps can discover different solutions (i.e., among different homotopic classes) and easily handle task-specific constraints (i.e., collisions, limits), but can be negatively impacted computationally as the cost of \sbmp subroutines increases, e.g., forward propagation, collision checking, and nearest neighbor discovery \cite{lavalle2006planning}, which is an issue when planning for dynamical systems.

Motion planning can also be viewed as a numerical optimization problem. Example optimization-based motion planners include Covariant Hamiltonian optimization (CHOMP)\cite{ratliff2009chomp}, stochastic trajectory optimization (STOMP) \cite{kalakrishnan2011stomp}, and Sequential Convex Optimization (TrajOpt) \cite{schulman2013finding}. Starting from an initial, potentially infeasible, \textit{guess}, an optimizer attempts to minimize an objective function subject to constraints, such as system dynamics, collisions, reaching the goal, and possibly other aspects, e.g., energy minimization. Optimization methods can quickly converge to a feasible, high-quality solution if the initial guess is in the vicinity of one. Non-trivial environments and systems, however, can challenge convergence and may require careful definition of parameters, such as \textit{obstacle potentials} \cite{ratliff2009chomp}. Frequently, the initial guess also imposes a discretization of the solution trajectory, limiting the set of possible solutions. Continuous trajectory representations via Gaussian Processes \cite{mukadam2018continuous} can minimize the discretization problem for linear time-varying (LTV) systems that are stabilized using stochastic differential equations (SDEs), i.e., LTV-SDE systems. 

There are also integrative frameworks in the planning literature. For kinematic systems, a precomputed graph of convex sets can be used by a mixed-integer optimizer to find a valid path \cite{marcucci2023motion}. An approach for dynamical systems involves the pre-computation of motion primitives, which are \textit{stitched} together by a planner \cite{honig2022db, ortiz2024idb}. An interleaving approach uses a graph to generate suggestions used by an optimizer \cite{natarajan2021interleaving}. Simultaneous localization and planning (SLAP) \cite{agha2018slap} models the challenge as a POMDP, which is approximated offline by a roadmap in belief space that is updated online via observations.

In the area of {\bf state estimation}, there has been a transition from tools that explicitly estimate the probability distribution regarding the robot's state to optimization techniques, such as those that employ Factor Graphs  (\fgs) \cite{dellaert2017factor}. A \fg is a probabilistic graphical model that can be used to represent the joint probability mass function of the variables that comprise the system \cite{KschischangFG}. \fgs aim to exploit the underlying, known structure of the problem by using sparse linear algebra techniques and find solutions via optimization algorithms, such as Gauss-Newton. Non-linear problems require frequent re-linearization, which is a costly operation to perform online over the entire graph. The incremental smoothing and mapping (iSAM2) \cite{kaess2012isam2} algorithm uses the Bayes tree \cite{kaess2011bayes} to keep updates local, avoiding a full re-linearization of the graph. \fgs are well suited for addressing a variety of problem formulations, such as simultaneous localization and mapping (SLAM) \cite{cunningham2010ddf,forster2015imu,jiang2021probabilistic} and simultaneous trajectory estimation and mapping (STEAM) \cite{barfoot2014batch}. Multiple initial guesses are possible via a network of trajectories so as to initialize a \fg -based planning solution for holonomic systems \cite{huang2017motion}. 

The two closest methods to this work are also employing \fgs and {\bf integrate state estimation and planning/control}. They correspond to Simultaneous Trajectory Estimation And Planning (\steap) \cite{mukadam2019steap} and Simultaneous Control And Trajectory Estimation (\scate) \cite{king2022simultaneous}. \scate builds on top of \steap and also deals with dynamic obstacles. Specifically, \steap employs Gaussian processes as dynamics, while \scate uses a linear time-invariant (LTI) approximation of the dynamics. \scate adds the LTI dynamics and dynamic obstacle factors to the \fg. The output of \steap is a \textit{plan} but does not compute controls directly and relies on an external controller. \scate outputs a control to be applied to the robot. To operate at high frequencies, \scate relies on a low-level controller and state estimator. Neither of the two techniques is directly applicable to systems with non-linear dynamics. Prior work also uses naive initialization of these methods, such as a straight line connection of the start and the goal, which can lead to local minima due to the local nature of numerical optimization.

\section{PROBLEM SETUP} \label{sec:preliminaries}

Consider a robot with state space $\mathbb{X}$ and control space $\mathbb{U}$ tasked with navigating a workspace $\mathbb{W}$ from initial state $x_0$ to a goal region $X_G$. Obstacles divide $\mathbb{X}$ into collision-free $\mathbb{X}_\mathrm{f}$ and obstacle $\mathbb{X}_\mathrm{o}$ subsets. The {\bf true dynamics} $\dot{x}_t = f(x_t,u_t)$ ($x_t \in \mathbb{X}$, $u_t \in \mathbb{U}$) govern the robot's motions but are not perfectly known and may exhibit non-holonomic constraints.

In terms of its dynamics, the robot has access only to an {\bf approximate dynamics model} via a parameterized function $\dot{x}_t = \hat{f}_\rho(x_t,u_t)$, which is defined by a set of physical parameters $\rho$. The approximate dynamics model $\hat{f}_\rho$ is a simplification of the true dynamics $f$ and does not necessarily have the same expression as $f$, i.e., there may not be a choice of physical parameters $\rho$, which will allow $\hat{f}_\rho$ to be perfectly identified with $f$. Example physical parameters for a car-like robot are friction coefficients, steering, and throttle gains.

In addition, the approximate model $\hat{f}_{\rho}$ employed models the obstacle-free workspace as a flat, planar surface with uniform friction.  In reality, however, the true workspace can also exhibit: (i) different friction from the one the robot assumes, which can vary over the workspace, and (ii) an uneven surface that can include traversable obstructions, such as a ramp.

A {\bf plan} $p_T$ is a sequence of $T$ piece-wise constant controls $\{u_0, \ldots, u_{T-1}) \}$, where each control $u_i$ is executed for a timestep $\Delta t_i$. When a plan $p_T$ is executed at a state $x_i$, it produces a {\bf trajectory} $\tau_f(x_t, p_T)$, i.e., a sequence of states $ \{x_t, \cdots, x_{t+T}\}$ according to a dynamics model $f$. Due to the gap between the true dynamics $f$ and the planning model $\hat{f}_\rho$, the executed robot trajectory $\tau_f(x_i, p_T)$ does not match the planned trajectory $\tau_{\hat{f}_\rho}(x_i, p_T)$ for the same plan $p_T$.

The robot has access to noisy sensing that provides discrete {\bf measurements} $z(t)$, which partially inform about the robot's state $x(t)$, such as sensing the robot's pose from external sensors. A state estimation process uses measurements $z(0:t)$ to compute {\bf estimated states} $\bar{x}(0:t)$. This work assumes perfect knowledge of obstacles' poses during execution, such as walls or obstacles that the robot should not collide with. 

A {\bf controller} $\pi_{\hat{f}}( \bar{x}_i, \tau_{\hat{f}_\rho} )$ is employed to track the planned trajectory $\tau_{\hat{f}_\rho}$ given the estimated states $\bar{x}_i$ and returns controls $u \in \mathbb{U}$.  With some abuse of notation\footnote{Trajectories above were defined for open-loop plans $p_T$. Here the definition is adapted to receive as input the controls arising from the closed-loop controller $\pi_{\hat{f}}$.}, denote as $\tau_f( x_i, \pi_{\hat{f}})$ the trajectory executed by the robot when the controller $\pi_{\hat{f}}$ is executed starting at state $x_i$.

{\bf Problem Definition:} Given a: (i) start state $x_0 \in \mathbb{X}_\mathrm{f}$, (ii) goal region $X_G\subset \mathbb{X}_\mathrm{f}$,  (iii) approximate dynamics model $\hat{f}_{\rho}(x,u)$ given identified parameters $\rho$, (iv) desired trajectory $\tau_{\hat{f}_\rho}(x_0, p_T)$ that brings the robot in $X_G$, and (v) online measurements $z(t)$, the objective is to simultaneously compute estimated robot states $\bar{x}(t)$ and execute a controller $\pi_{\hat{f}}( \bar{x}_0, \tau_{\hat{f}_\rho} )$ so that the executed trajectory $\tau_f(x_0,\pi_{\hat{f}})$ is collision free and brings the robot inside $X_G$. 

Secondary objectives include minimizing the error between estimated states $\bar{x}(t)$ and true states $x(t)$, minimizing the error between the planned and the executed trajectory, as well as minimizing the cost of the executed trajectory. In this work, the cost corresponds to the trajectory duration. 

{\bf Additional notation: }
The goal region is defined by a single configuration $q_G$ in the robot's configuration space $\mathbb{Q}$ so that: $X_G = \{x \in \mathbb{X}_f \ \vert \ d(\mathbb{M}(x), q_G) < \epsilon \}$, or equivalently, $X_G = \mathcal{B}(q_G, \epsilon)$, where $\epsilon$ is a goal radius in  $\mathbb{Q}$ according to distance function $d$. The function  $\mathbb{M}(x)$ maps states to configurations. Estimated \textit{past}/\textit{current} states are represented by $\bar{x}(t)$ and future (estimated) states are represented as $\hat{x}(t)$.

The following are relevant Lie group concepts and operations; refer to \cite{sola2018micro} for a more in-depth explanation.  Consider $q_i \in \mathcal{M}$ where $\mathcal{M}$ is a Lie group of dimension $m$ and $\dot{q}_i \in \mathbb{R}^m ( \cong T_{q_i}\mathcal{M} )$ is a constant velocity (an element in the tangent space of $\mathcal{M}$ at $q$). 
The function $ \texttt{Exp} :  \mathbb{R}^m \rightarrow \mathcal{M} $ maps vector elements to the manifold with its inverse being $ \texttt{Log} : \mathcal{M} \rightarrow \mathbb{R}^m$. 
The $\texttt{Between}: \mathcal{M} \rightarrow \mathcal{M}$ operation is defined as $\texttt{Between}(q_a,q_b) = q_a^{-1} \circ q_b$ and computes the element that would \textit{move} $q_a$ to $q_b$. 
Forward integration in a Lie group is defined as $ q_{i+1} = q_i \circ \texttt{Exp}( \dot{q}_i \Delta t_i)$.


\begin{figure}[t]
    \centering
		\includegraphics[width=0.45\textwidth]{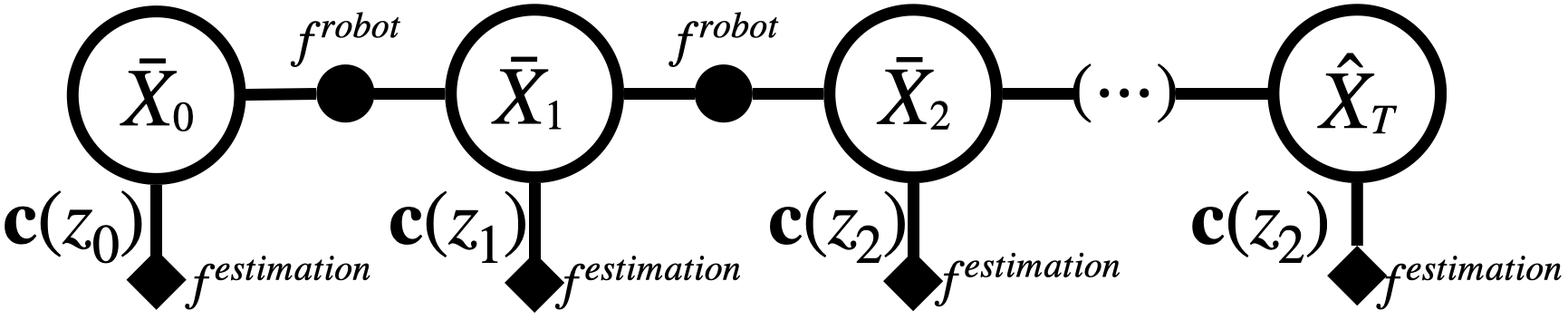}
    \caption{\small A typical trajectory estimation \fg at time $T$ uses state observations $z^x(0:T)$ and the robot model $\dot{x}_t = \hat{f}_\rho(x_t,u_t)$ to generate state estimates $\bar{X}(0:T)$. The unary factors impose a cost between observations and estimated states. The binary factors correspond to the robot's dynamics.}
    \label{fig:estimation_fg}
\vspace{-0.2in}
\end{figure}

\section{Foundations: Inference via Factor graphs}
\label{sec:factor_grapsh}

Factor graphs (\fgs) have been highly adopted for sensor fusion, state estimation, and localization problems. Given the relationship between estimation and control, recent work also explores their use in control and planning \cite{barfoot2014batch, mukadam2019steap, mukadam2018continuous}. 

In probabilistic inference, the objective is to find the set of values $\theta$ given events $e$. The posterior density of $\theta$ is computed via Bayes rule: $p(\theta|e) \propto p(\theta)p(e|\theta)$, where $p(\theta)$ is the prior on $\theta$ and $p(e|\theta)$ is the likelihood function. Given a prior and a likelihood, the optimal solution is found by the \textit{maximum a posteriori} (MAP) operation:
\vspace{-0.05in}
$$ \theta^* = \arg\max_\theta p(\theta|e) = \arg\max_\theta p(\theta) l(\theta;e),\vspace{-0.05in}$$
where the likelihood function  $l(\theta;e)= p(e|\theta)$ specifies the probability of events $e$ given $\theta$. The general likelihood function for non-linear \fgs is defined as: $l(\theta;e) \propto \exp(\frac{1}{2}|| h(\theta, e)||^2_{\Sigma})$, where $h$ is a \textit{measurement function} with covariance $\Sigma$.   

Formally, a \fg is a bipartite graph $FG = (\Theta,\mathcal{F},\mathcal{E})$ with two types of nodes: factors $f_i \in \mathcal{F}$ and variables $\theta_i \in \Theta$. \fg edges are always between factor nodes and variable nodes. Given an \fg, its posterior distribution is:
\vspace{-0.1in}
$$ p(\theta|e)\propto \prod_{n=0}^N f_n (\theta_n).
\vspace{-0.1in}$$
The MAP estimate can be reduced to a non-linear least squares problem and solved with standard solvers. Most robotic problems require the solver to operate at high frequency and incorporate new data on demand.  Standard solvers are not enough in robotics, as they do not take advantage of the sparsity or the \textit{incremental} nature of robotic problems. To alleviate this, iSAM \cite{kaess2008isam} exploits the problem's structure imposed via the \fg. iSAM uses a Bayes tree to avoid re-linearization of variables unaffected by a new measurement. Finally, a \fg implies some level of discretization, which is also present in other estimation and trajectory optimization approaches. The following subsections review \fg representation from the literature for estimation and planning/control.

\subsection{Trajectory Estimation}
\label{sec:trajest}

Fig.~\ref{fig:estimation_fg} presents a typical \fg for past trajectory estimation. It computes  $p(\theta_{TE}|e)\propto f^{TE}=f^{traj}f^{estimation}$ where $f^{traj}=\prod_{i=0}^{t}f^{robot}_i(\theta_i)$ is the trajectory derived from the robot's model and $f^{estimation}=\prod_{i=0}^{t}f^{estimation}_i(\theta_i) $ are factors that incorporate the measurements. Typically, the underlying discretization is given by the frequency of the measurements. Discretization can be problematic for complex dynamical systems. To avoid discretization of the dynamics, methods often employ a Gaussian Process (GP) to model the system, achieving continuous-time trajectories but limiting applicability to holonomic robots with linear expressions. 

\textbf{Benefits of Trajectory Estimation vs Filtering} While a controller only requires the latest state, methods such as Kalman/particle filters, which provide incremental estimates of the latest state, solve the Bayesian \emph{filtering} problem. The latest robot state estimate, however, of an optimal solver for the \emph{most likely trajectory} problem, which \stela focuses on, can be different (and more accurate/less uncertain) than that of an optimal solver for \emph{filtering}. Solving for the \emph{most likely trajectory} is typically computationally more expensive than solving \emph{filtering}. Given the least square approximations, factor graph frameworks allow solving such problems online. \stela takes advantage of this to provide high-frequency MLE updates of the robot's past trajectory. In this way, it provides better estimates of the robot's latest state, and future controls are simultaneously co-optimized based on these improved state predictions.

\subsection{Trajectory Optimization as a Motion Planner}
\label{sec:trajopt}

Motion planning can be seen as an optimization problem where the cost of the trajectory ${\bf cost}(\tau)$ produced by the plan $p_T$ is minimized subject to i) start state condition: $\tau(0) = x_0$, ii) goal condition: $\tau(T) \in  X_G$, iii) dynamics condition: $\dot{x}_t = \hat{f}_\rho(\hat{x_t},u_t)$ and iv) collision-free condition $\tau_{\hat{f}_\rho} \in \mathbb{X}_\mathrm{f}$. 
Additionally, problem-specific constraints can be added, i.e., smoothness on the controls, energy minimization, etc. 

\begin{figure}[t]
    \centering
		\includegraphics[width=0.45\textwidth]{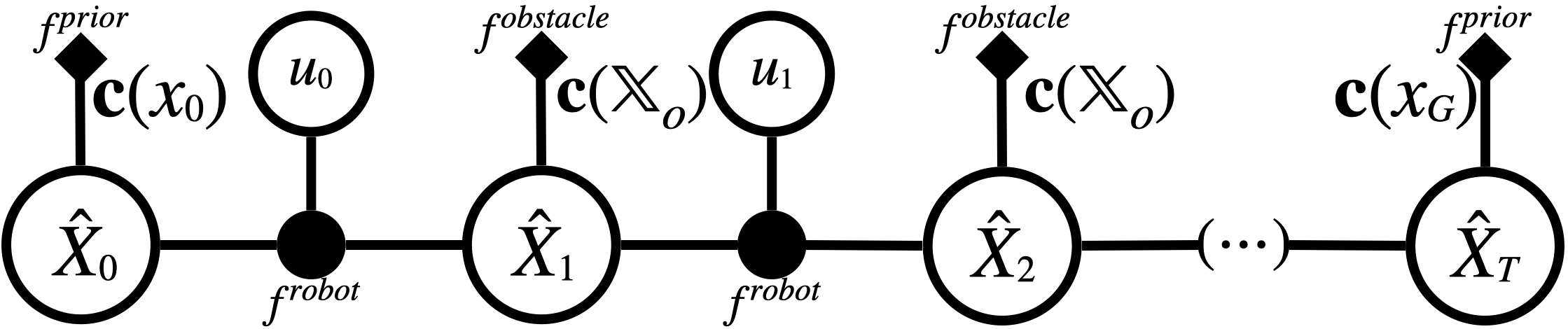}
    \caption{\small An \fg for robot planning employs the robot's model $\dot{x}_t = \hat{f}_\rho(\hat{x_t},u_t)$ on a dynamics factor to compute a trajectory of $T$ states, starting in $\hat{x}_0$ and ending in the goal region $\bar{X}_T \in X_G$. Beyond the ternary dynamics factor, there are costs imposed for the optimization by unary factors for obstacle avoidance  (${\bf c}(\mathbb{X}_o)$) over the intermediate state variables $(\hat{X}_1:\hat{X}_{T-1})$, a state prior for the initial state (${\bf c}(x_0)$) and a goal region prior for the final state (${\bf c}(X_G)$).}
    \label{fig:planning_fg}
    \vspace{-0.1in}
\end{figure}

\fgs have also been used for motion planning of holonomic systems.  In the factor graph setting, trajectories can be obtained via: $p(\theta_{Pl}|e)\propto f^{robot}f^{obstacle}f^{prior}$ \cite{mukadam2018continuous}, where $f^{robot}$ are the dynamical system factors, $f^{obstacle}$ the obstacle avoidance factors and $f^{prior}$  the (fixed) start and goal conditions. Fig.~\ref{fig:planning_fg} provides an example.

For a general dynamical system, each factor $f^{robot}$ corresponds to solving a \textit{steering function}. Dynamics linearization has been proposed as an alternative to solving a steering function for certain systems \cite{webb2013kinodynamic,goretkin2013optimal}. In practice, finding a feasible solution depends both on the quality of the prior and the complexity of the environment. In previous factor graph approaches, these limitations are alleviated by using a holonomic robot modeled via a Gaussian Process and relying on random re-initializations.

\fgs can suffer from local minima, making any solution heavily dependent on the \textit{initial guess}. For the motion planning problem, a good initialization is not always trivial: a dynamically feasible solution may be in collision while a collision-free one may not be feasible.

\section{Simultaneous Trajectory Estimation\\ and Local Adaptation (\stela)}
\label{sec:method}

\begin{figure}[t]
    \centering
    \includegraphics[width=\linewidth]{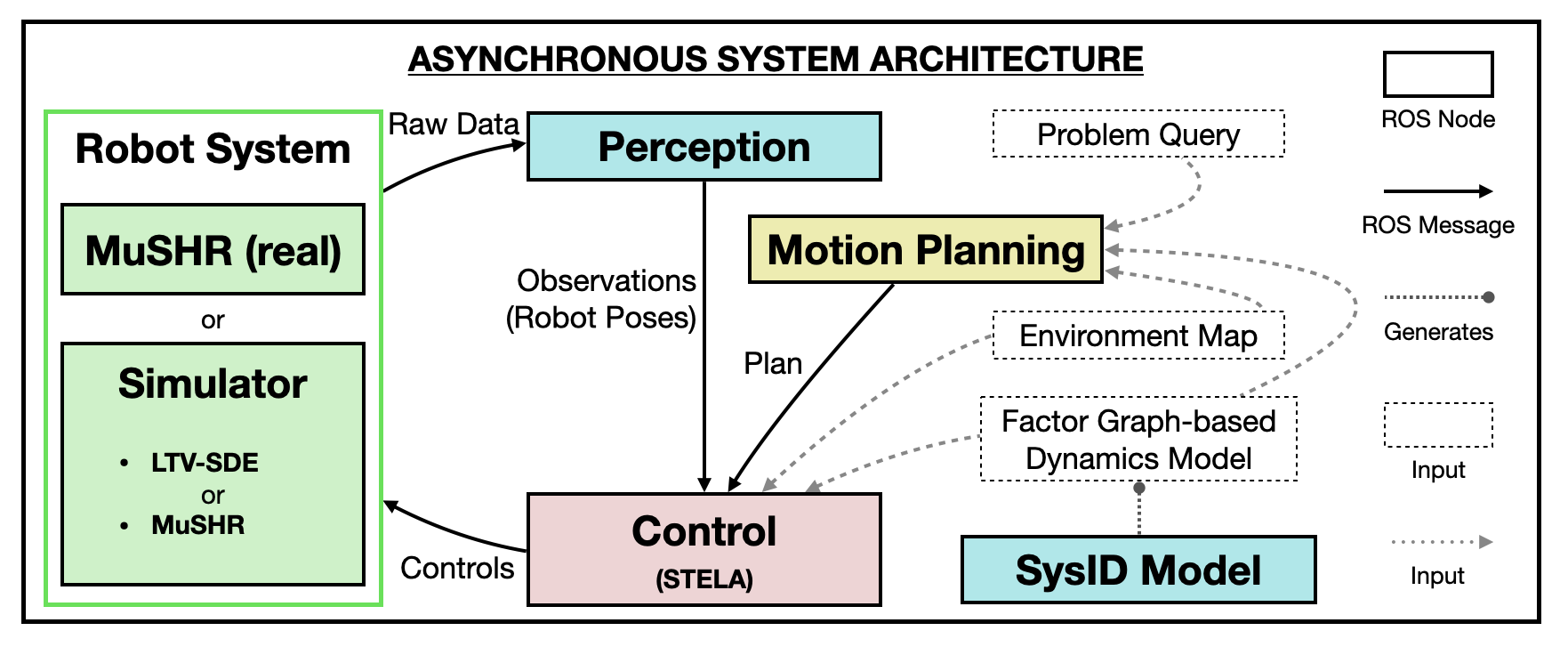}
    \vspace{-0.2in}
    \caption{\small Asynchronous system architecture: Offline, a system identification process generates a \fg-based dynamics model of the robot system. A motion planner receives the dynamics model, the environment, and a motion planning query as input to generate a desired solution plan that addresses the query for the given map and model.  Upon completion, the desired plan, the model, and the environment are sent to \stela. Online, raw data --i.e., images from external cameras-- are processed by a perception process to provide robot state observations to the control module. These observations are used by \stela to estimate the executed trajectory and generate controls to be forwarded to the robot at a high frequency. The closed-loop framework enables the system to adapt to noise dynamically, execution errors, and the gap between the planning model and the real system. In the accompanying experiments, the robot system is either a real \mushr robot or a simulated system, where both an idealized LTV-SDE robot and an analytical dynamics model of a \mushr robot are considered.}
    \label{fig:stela-sys-arch}
    \vspace{-0.15in}
\end{figure}

The \stela framework uses numerical optimization for simultaneously solving trajectory estimation and control. It introduces a general \fg representation tailored for kinodynamic trajectory following, given feasible desired plans generated by an \sbmp. Fig.~\ref{fig:stela-sys-arch} presents the overarching system and the processes it involves. Offline, a system identification process builds a dynamics model $\hat{f}_\rho(x_t,u_t)$ that bridges the gap with the target robot (see the Experiments Section for this process, which is based on  \fg tools). Given the model, a kinodynamic \sbmp is called to solve a motion planning query given the available environment map. The resulting feasible plan $p_T$ from the planner is forwarded to the \stela module, which also consumes from a perception system online observations $z(t)$ regarding the robot's poses. \stela internally generates improved estimates $\bar{x}(0:t)$ of past robot states and forwards controls $u(t)$ to the robot that minimize deviation from the planned trajectory, move the robot to the desired goal region $X_G$, and avoid collisions.

\subsection{Initialization of desired trajectory via \sbmp}

Given the identified robot model $\hat{f}_\rho(x_t,u_t)$, an environment map that identifies obstacle regions $\mathbb{X}_\mathrm{o}$ and a motion planning query specifying $x_0$ and $X_G$, the approach calls an asymptotically optimal kinodynamic Sampling-Based Motion Planner (\sbmp) \cite{kleinbort2020refined, littlefield2018efficient} tasked to generate a feasible, collision-free trajectory $\tau_{\hat{f}_\rho}(x_0, p_T)$ that solves the query on the provided map for the given model.

The output of the planner is treated as the desired trajectory. It is represented via a discretized graphical representation $\mathcal{G}=(N,E)$ where a node $n_i \in N$ represents a reachable, collision-free state $x_{i} = [q_i, \dot{q}_i]^T$  and an edge $e_{ij} = \{n_i,n_j\} \in E$ contains a control, duration pair $(u,\Delta t)$ that drives the robot from state $n_i$ to state $n_j$ according to $\hat{f}_\rho$.   An upper threshold for the duration $\Delta t$ of edges $e_{ij} \in E$ of the desired trajectory is applied (0.5sec for the LTV-SDE system and 0.1sec for MuSHR in the accompanying experiments). If a control is applied for longer than $\Delta t$ in the solution \sbmp trajectory, then it is broken into multiple smaller edges in the graphical representation of the desired trajectory so that none exceed the duration threshold. In this way, the number of discrete states used to initialize \stela is not fixed, and it is adaptive to the output solution of the \sbmp.  

\subsection{The \stela Factor Graph}

\begin{figure*}[thpb]
    \centering
    \begin{subfigure}{.42\textwidth}
        \includegraphics[width=0.95\linewidth]{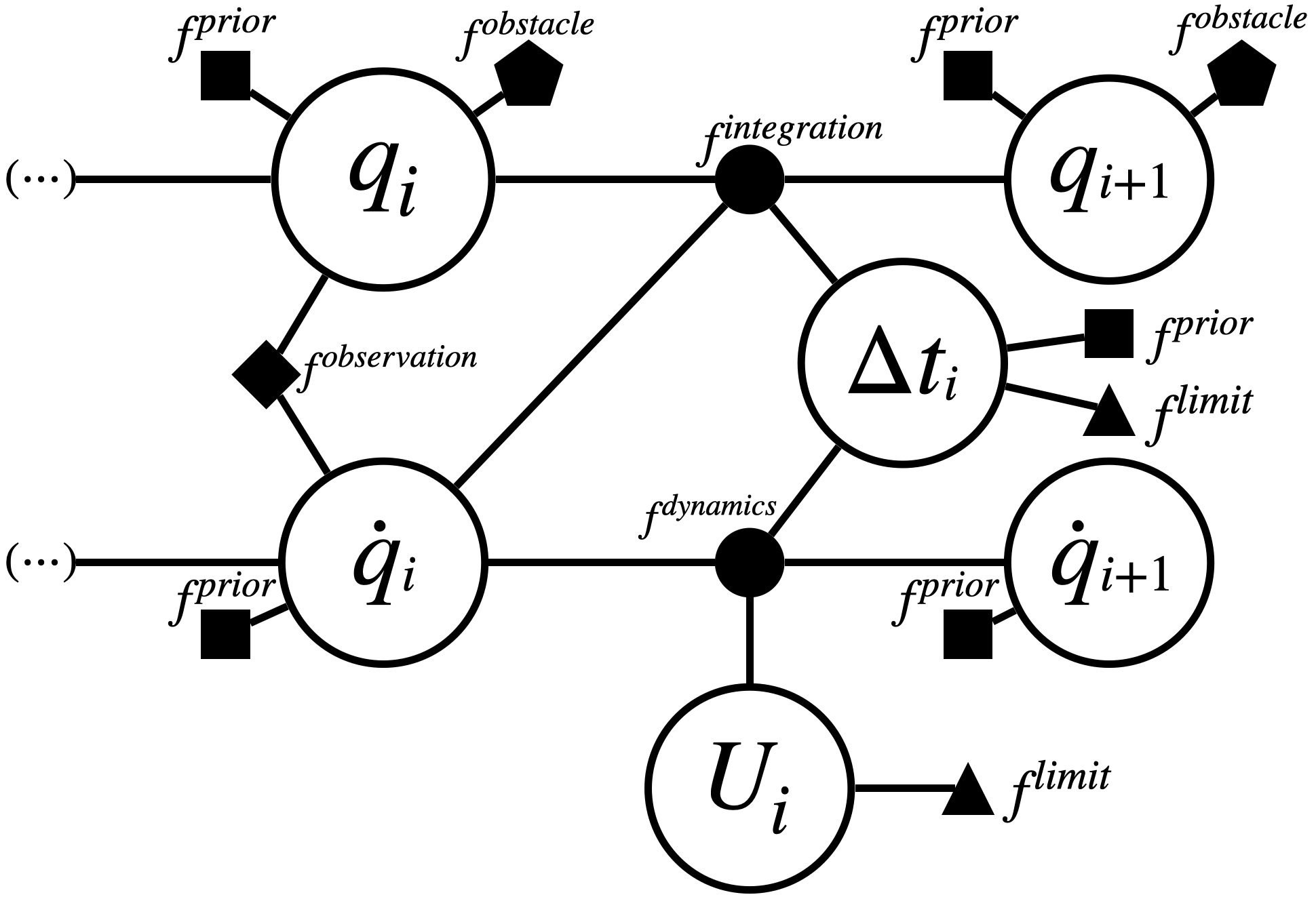}
        \caption{Dynamics Factor Graph}
        \label{fig:dynamics_fg_explicit}
    \end{subfigure}%
    \begin{subfigure}{.15\textwidth}
        \includegraphics[width=0.95\linewidth]{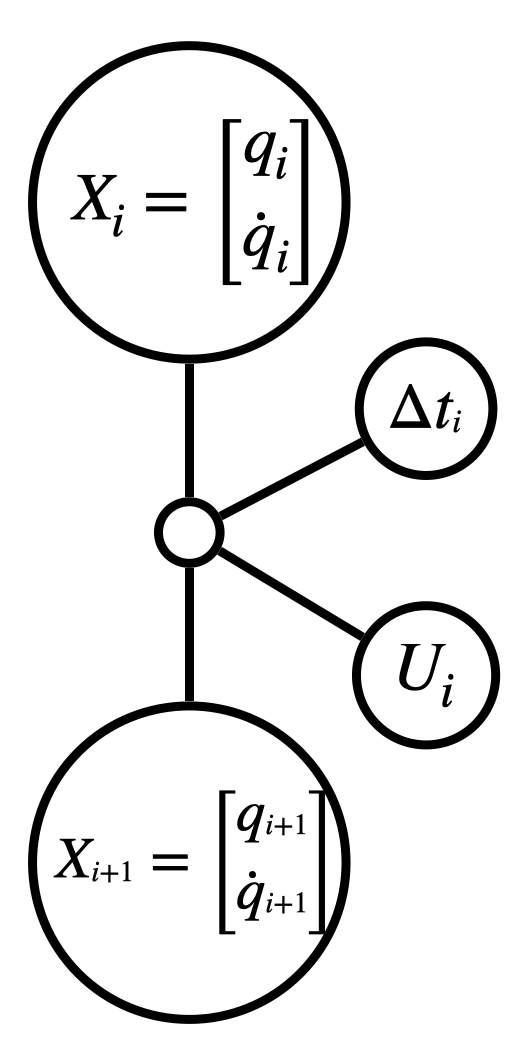}
        \caption{\small Compressed Dynamics Factor Graph}
        \label{fig:dynamics_fg_compress}
    \end{subfigure}%
    \begin{subfigure}{.42\textwidth}
        \includegraphics[width=0.95\linewidth]{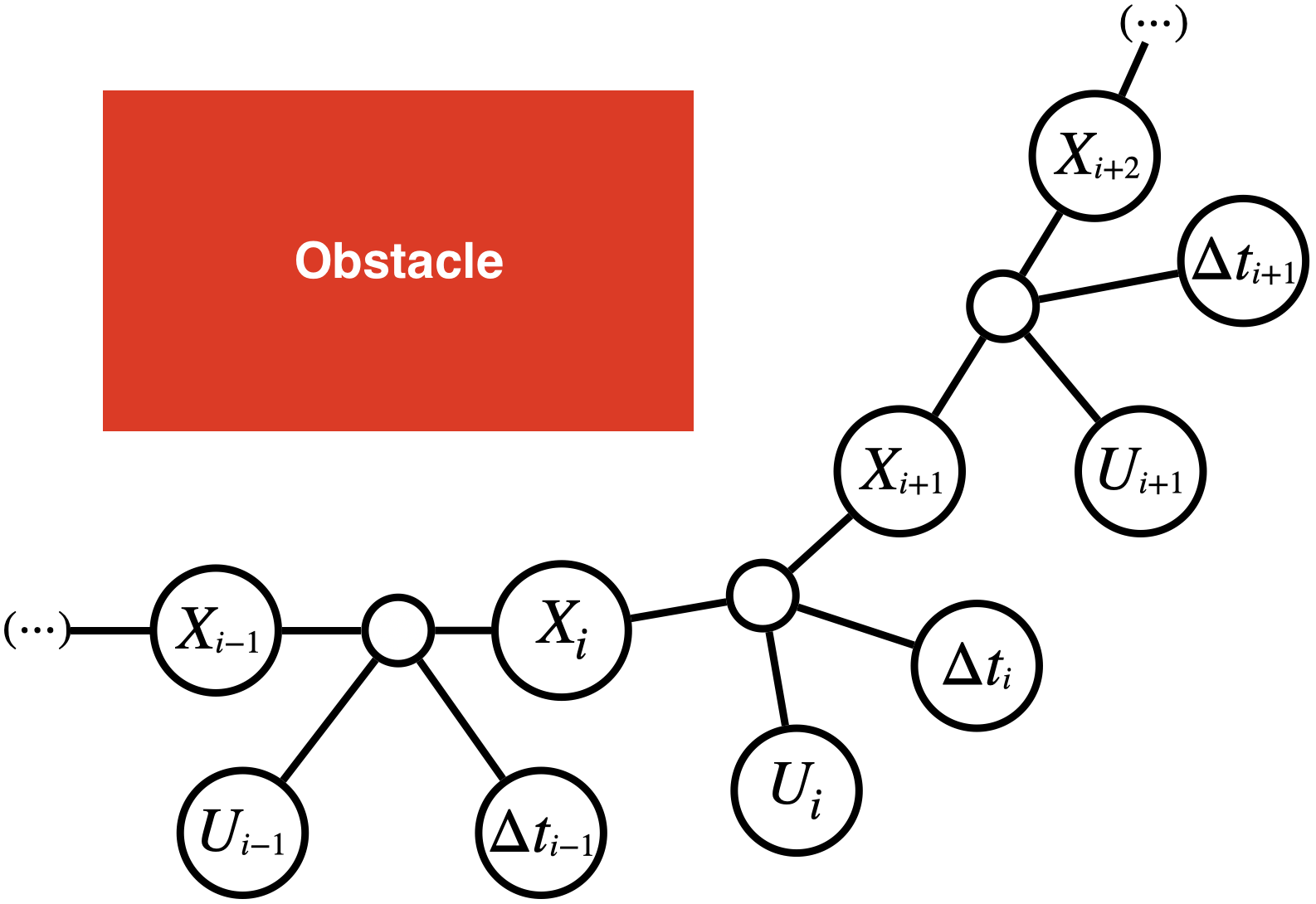}
        \caption{\stela Trajectory}
        \label{fig:stela_traj_compress}
    \end{subfigure}%
    \caption{\small (Left) The dynamics factor graph corresponding to each edge of the desired trajectory with all associated factors. (Middle) For visualization purposes, the dynamics factor graph is also presented in a \textit{compressed} form, which is symbolized by a hollow factor. (Right) A collision-free trajectory, consisting of a dynamic factor graph sequence, is shown.}
    \label{fig:dynamics_fg}
    \vspace{-0.15in}
\end{figure*}

\stela builds on top of the incremental smoothing and mapping (iSAM) framework \cite{kaess2008isam} and is executed at a high frequency to consume robot measurements $z(t)$ that arrive asynchronously. \stela is initialized by converting each edge of the desired trajectory into a \emph{dynamics factor graph} as in Fig.~\ref{fig:dynamics_fg}. The proposed \fg includes six different factor types. Each factor is defined as $f^{j}(\cdot) \propto \exp(\frac{1}{2}||h^j(\cdot)||^2_{\Sigma})$ for a factor-specific error function $h^j(\cdot)$. All the \fg variables  $q_i, \dot{q}_i,$ $u_i$ and $\Delta t_i$ are initialized according to the desired trajectory. 

The \textbf{integration factor} operates a configuration $q_i$, the velocity $\dot{q}_i$, the duration $\Delta t_i$, and the predicted next configuration $q^{pred}_{i+1} = q_i \circ \texttt{Exp}( \dot{q}_i \Delta t_i)$. The error function is then defined as $h^{integration}(q_{i+1},q_i,\dot{q}_i, \Delta t_i) = \texttt{Log}(\texttt{Between}(q^{pred}_{i+1},q_{i+1} ))$.

The \textbf{dynamics factor} explains the evolution of the velocity given the control input. The control input $u_i$ is used to obtain an acceleration in the local frame via a system-specific function $\ddot{q}=f(u_i)$.  The error function uses the predicted velocity term $\dot{q}^{pred}_{i+1} = \dot{q}_i + \ddot{q} \Delta t$ to compute the error function: $h^{dynamics}(\dot{q}_{i+1}, \dot{q}_{i}, u_i, \Delta t_i) = \dot{q}^{pred}_{i+1} - \dot{q}_{i+1}$. 
As $\dot{q} \in \mathbb{R}^m$, an Euler integration scheme is sufficient.

The \textbf{observation factor} incorporates observations to estimate the executed trajectory. This work considers observations of the configuration $q^i_z$ that are generated asynchronously as the robot moves. Observations have a known (but noisy) timestamp from which the elapsed time from $q_i$ to the observation is $\Delta t^i_z$. The predicted observation is then $q^{pred}_{z} = q_i \circ \texttt{Exp}( \dot{q}_i \Delta t_z)$ and the error is $h^{observation}(q_i,\dot{q}_i) = \texttt{Log}(\texttt{Between}(q^{pred}_{z}, q_{z}))$. Observation factors using measurements of velocities or higher-order magnitudes can be integrated in a similar scheme. 

The \textbf{prior factor} is a unary factor that penalizes deviations of some factor graph variable $v$ from a given constant value $v^{prior}$. The error is defined as: $f^{prior}(v) = \texttt{Log}(\texttt{Between}(v, v^{prior}))$. Prior factors are added to $q_i, \dot{q}_i,$ and $\Delta t_i$ variables given the desired trajectory. They are not applied, however, to control $u_i$ variables to provide the ability to \stela to adapt the future trajectory given the latest state estimates. Control variables are only initialized to the value corresponding to the desired trajectory.

The \textbf{obstacle factor} introduces a notion of safety by \textit{pushing} states away from obstacles. While the \sbmp trajectory initialization is collision-free, a robot may move dangerously close to obstacles due to the model gap and noise. Th factor, following the definition in \cite{mukadam2018continuous}, uses a distance function $d(\cdot)$ to obstacles (from the environment map) and an $\epsilon$ threshold: \vspace{-.05in}$$h^{obstacle}(q_i) = \begin{cases} -d(q_i) + \epsilon, & d(q_i) \leq \epsilon; \\ 0, & d(q_i) > \epsilon.\end{cases}\vspace{-.05in}$$ 

Two obstacle factors are considered: one using a precomputed SDF and another using distance computations from a collision checker, i.e., the PQP library \cite{larsen1999fast}. The SDF factor allows for precomputation of the environment and only considers the distance to the closest obstacle. The PQP factor can be called online and is defined per obstacle, allowing multiple factors to be \textit{active} at the cost of increased computation. Given a reasonable threshold distance $\epsilon$,  even in a cluttered environment, most obstacle factors will remain \textit{inactive}. As the error is zero in these cases, re-linearization is unnecessary.

The \textbf{limits factor} penalizes values that exceed a predefined value $v^{lim}$. The error for an upper limit of a variable $v$ is: 
$$h^{limit}(v) = \begin{cases} v - v^{lim}, & v \geq v^{lim}; \\ 0, & \texttt{otherwise.}\end{cases}$$ 
Lower limits can be computed similarly. Multi-value limits are defined element-wise. Limit factors are applied to control variables $u_i$ to guarantee a feasible solution and the duration variables $\Delta t_i$ to keep them within a reasonable range.

Fig.~\ref{fig:dynamics_fg_explicit} shows the dynamics factor graph constructed from two \sbmp-nodes and one \sbmp-edge. Each \sbmp-edge introduces one integration factor and one dynamics factor. Obstacle factors are added per configuration. Limit factors constraint $\Delta t_i > 0 $ and controls as the \sbmp solution must respect physical limits. Five prior factors are included, one per configuration $q$, one per velocity $\dot{q}$, and one for the duration $\Delta t_i$ to penalize deviations from the desired trajectory. Finally, observation factors are added as new observations arrive.

\subsection{Inference over a Sliding Window}
\begin{figure*}[thpb]
    \centering
	\centering
	\includegraphics[width=0.9\textwidth]{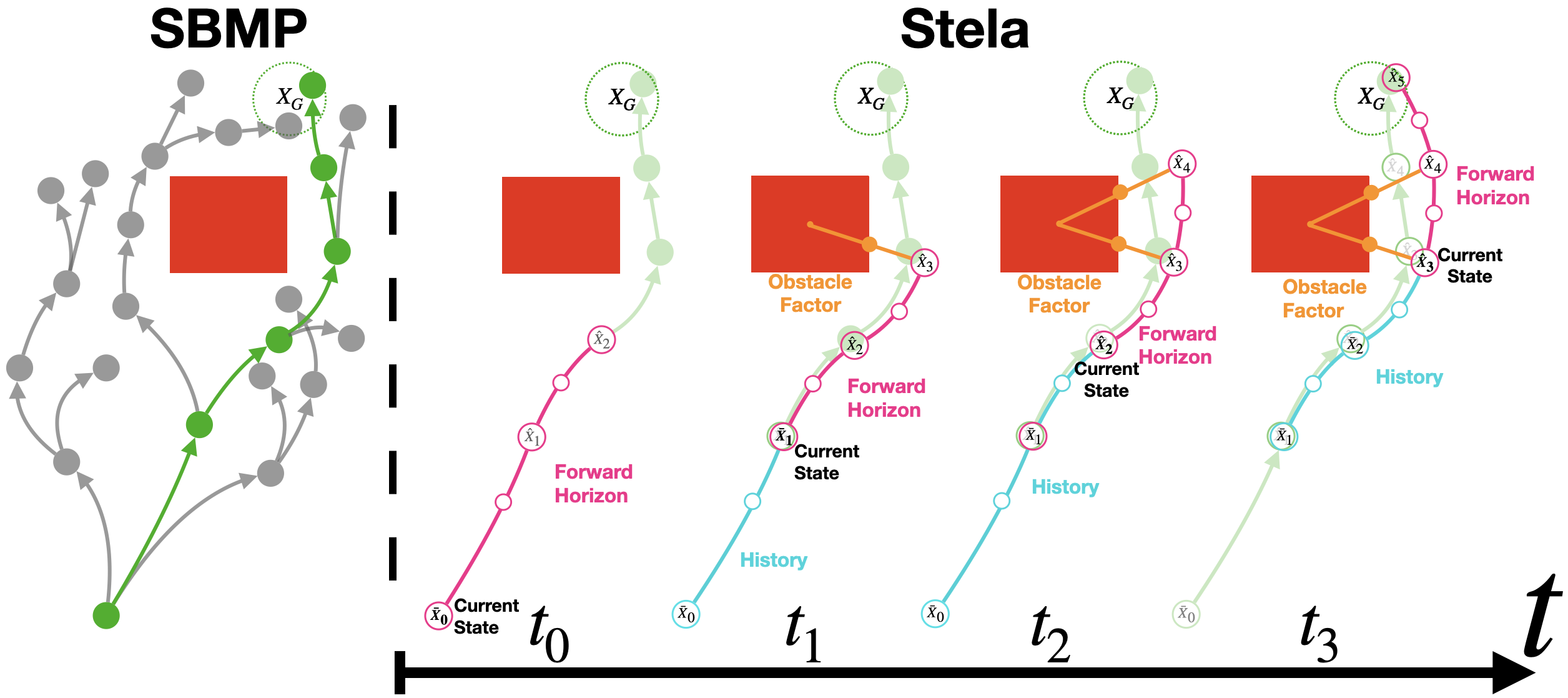}
	\caption{\small \stela with a sliding window in action: Offline, the SBMP generates a graphical representation from which a feasible, collision-free, goal-reaching trajectory is obtained. At initialization, the factor graph converts forward-horizon nodes (image: $n^{fwd}$ = $n^{hist}$ = 2 ) into a dynamics factor graph and starts execution. Running at a given frequency, at each iteration, \stela updates the factor graph, incorporating new observations and moving the window. The window update at time $t_i$ is determined by comparing the variable $\Delta t_i$ to the elapsed time since the last window update. Window updates incorporate the next node from the planned trajectory and (if necessary) delete history nodes.}
    \vspace{-.15in}
	\label{fig:stela_window}
\end{figure*}

A \fg associated with the entire desired trajectory returned by the \sbmp planner has at least $2|N|+2|E|$ \fg-variables and $6|N|+5|E|$ \fg-factors. At runtime, the number of factors can further and quickly increase due to high-frequency observations. A sliding window approach (Fig.~\ref{fig:stela_window}) is proposed to alleviate the resulting computational cost of the optimization. The window is divided into a forward horizon using $n^{fwd}$ future nodes of the planned trajectory relative to the current state and a limited past history of $n^{hist}$ past variables. 

For the past history, \stela follows similar strategies for trajectory estimation as in the literature \cite{mukadam2019steap} (i.e., related to Section \ref{sec:trajest} and Figure \ref{fig:estimation_fg}), so that $p(\theta_{TE}|e)\propto f^{TE}= \prod_{j=curr- n^{hist}}^{curr} f^{integration}_j f^{dynamics}_j f^{observation}_j$, where $curr$ is the current state. In contrast to previous approaches (Fig.~\ref{fig:estimation_fg}), \stela only estimates the past $n^{hist}$ states, and the forward propagation process is modeled by the combination of the general integration and dynamics factors of the proposed \fg. The prior, limit, and obstacle factors are not used for the trajectory estimation component of the optimization. 

For the forward horizon, \stela does not perform planning or control from scratch given a naive initialization as in Section \ref{sec:trajopt} and Fig.~\ref{fig:planning_fg}. Instead, \stela locally adapts the \sbmp plan. Local adaptation refers to adapting the controls to minimize the error with the desired trajectory, respect the dynamics, and avoid obstacles. Thus, the forward horizon part of the window performs a local adaptation given: $p(\theta_{LA}|e)\propto  \prod_{j=curr}^{curr+ n^{fwd}}f^{integration}_j f^{dynamics}_j f^{obstacle}_j f^{prior}_j f^{limits}_j$. The observation factors are not used for the local adaptation component of the optimization.

\stela simultaneously performs trajectory estimation and local adaptation through the inference of:
\begin{align}
p(\theta_{STELA}|e)\propto f^{TE}f^{LA}
\end{align}

\subsection{The \stela Algorithm}

\begin{algorithm}
\caption{\stela}\label{alg:stela}
\begin{algorithmic}[1]
\Require $\mathcal{T}_{sbmp}$
\State \fg $\gets$ sbmp\_to\_fg($\mathcal{T}_{sbmp}$)
\State curr $\gets$ 0
\While{$\bar{x}(t) \notin X_G$}
	\State $j \gets max\{curr - n^{hist},0\}$ \Comment{History}
    \State $k \gets min\{curr + n^{fwd}, length($\fg$)\}$ \Comment{Horizon}
	\For{$ i \in \{j, \dots, k\}$ } \Comment{\textit{Lookahead}}
    	\State $\hat{x}_i \gets \fg$.estimate($i$)
    	\State $S_i \gets \fg$.covariance($i$)
  	\EndFor 
    \State \fg += $f^{observation}(q_{curr}, \dot{q}_{curr}, z_{new})$ \Comment{Add $z_{new}$}
    \State $\hat{dt}_{curr} \gets $FG.estimate($\Delta t_{curr}$) 
    \State $\hat{u}_{curr} \gets $FG.estimate($u_{curr}$) 
    \Comment{Adapt control}
    \State send\_control($\hat{u}_{curr}$)
    \If{ (clock() - prev) $> \hat{dt}_{curr}$} 
        \State prev = clock() 
        \State \fg -= $\{(q_{j},\dot{q}_{j},u_j,\Delta t_j\}$  \Comment{Remove history} 
        \State \Comment{Add to Forward Horizon} 
        \State \fg += \{ $f^{integration}(q_{k+1}, q_{k},\dot{q}_{k+1}, \Delta t_{k})$,  $f^{dynamics}(\dot{q}_{k+1}, \dot{q}_{k},u_{k+1}, \Delta t_{k})$, $f^{prior}(q_{k+1}), f^{prior}(\dot{q}_{k+1}), f^{prior} (\Delta t_{k})$,  $f^{limits}(u_{k+1}), f^{limits}(\Delta t_{k})$, $f^{obstacle}(q_{k+1})$ \}
        \State curr++ \Comment{Move to next state}
	\EndIf 
 
\EndWhile
\end{algorithmic}
\end{algorithm}

At the beginning of the execution, the initial $n^{fwd}$ nodes of the desired trajectory $\mathcal{T}_{sbmp}$ are converted into a factor graph \fg. At each iteration, while the goal has not been reached, \stela performs the following operations: lookahead, trajectory estimation, addition of observations, and local adaptation. 

\stela seeks to exploit the incremental nature of the iSAM2 algorithm by locally updating and querying the factor graph as necessary. The lookahead step obtains the estimated states of the forward horizon alongside their associated error.  New observations are added to the corresponding state, which may change the estimation of $q_i$,$q_{i+1}$,$u_{i}$ or $\Delta t_i$; obtaining the current estimate for $u_i$ and $\Delta t_i$. When the estimated duration of the current edge $\Delta t_i$ elapses, \stela \emph{moves} to the next node by a) deleting the history variables and associated factors to keep $n^{hist}$ forward variables, b) adding the next node from the planned trajectory and c) changing the current node.




\subsection{\stela vs Previous Factor Graph-based Approaches} 
\stela can be seen as generalization of both \steap \cite{mukadam2019steap} and \scate \cite{king2022simultaneous} but has also unique features.

A. The formulations of \steap and \scate restrict their applicability to holonomic dynamical systems. In the case of \steap, the dynamics are modeled via a Gaussian Process. \stela's formulation allows the use of general dynamical systems expressed via first or second-order analytical expressions. 
    
B. The prior approaches collapse $q$ and $q_t$ into a single factor, but \stela introduces separate factors, which results in a better exploitation of the sparsity characteristics of \fgs.
    
C. The sliding window mechanism keeps the number of factors in the \fg relatively constant (up to the number of observation factors). This allows for predictable, high-frequency updates and better performance over long trajectories and complex environments. 

D. \stela utilizes Lie operations for the integration factor, enhancing the approximation of dynamics and accelerating performance. This also renders \stela broadly applicable to a wide range of robotic systems.
    
E. Planning via an optimization approach requires continuously solving a Boundary Value Problem ({\tt BVP}) (or access to a steering function). This can result in solutions with local minima or the inability to converge in systems with non-linear dynamics. A core insight behind \stela is that this challenge can be simplified if a feasible initiation from an \sbmp is used, where the \sbmp has explored the state space for high-quality, feasible, and collision-free solutions.

\begin{figure}[]
\centering
\begin{subfigure}{.16\textwidth}
  \centering
  \includegraphics[width=.95\linewidth]{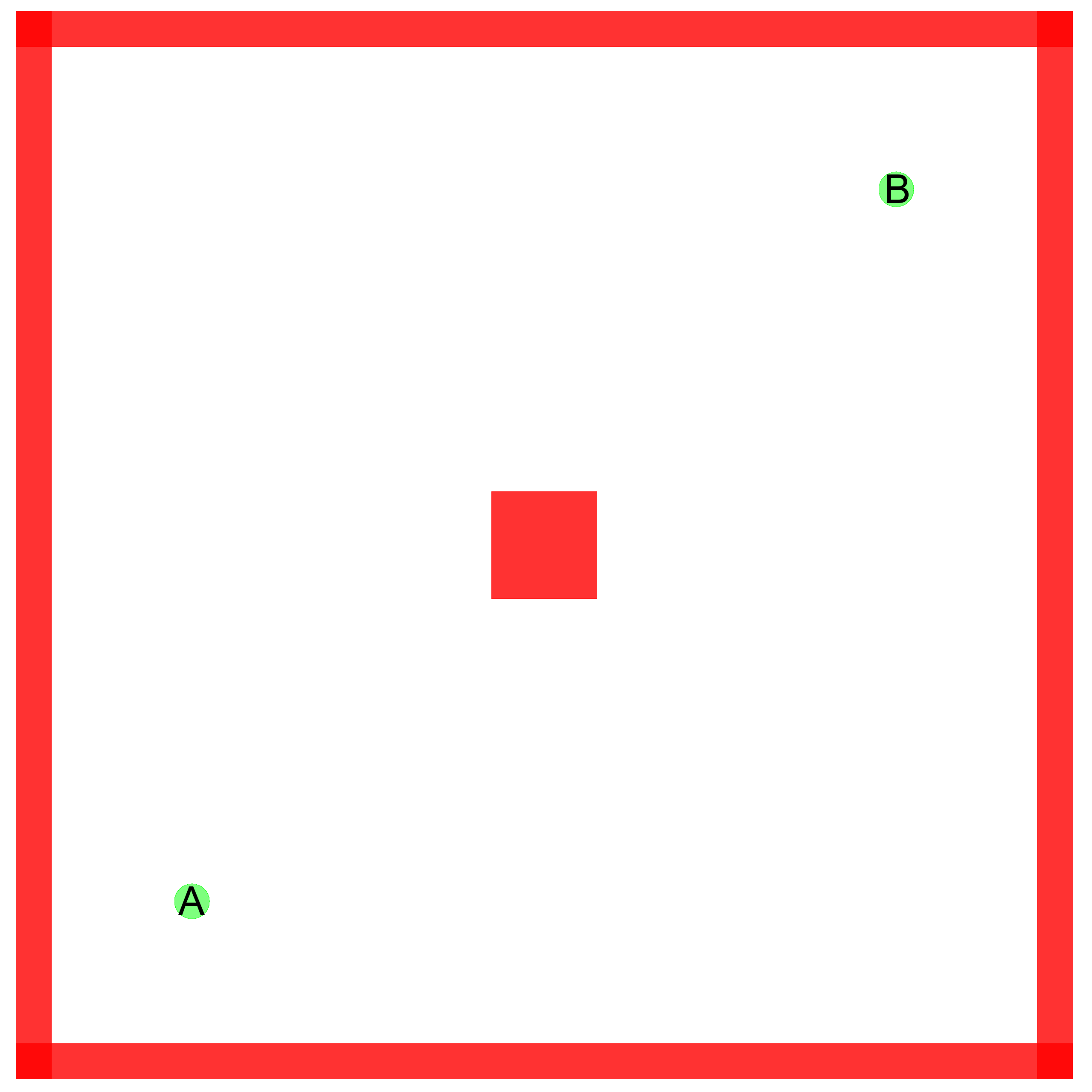}
  \caption{Simple Obstacle}
  \label{fig:empty_envs_simple_obstacle}
\end{subfigure}%
\begin{subfigure}{.16\textwidth}
  \centering
  \includegraphics[width=.95\linewidth]{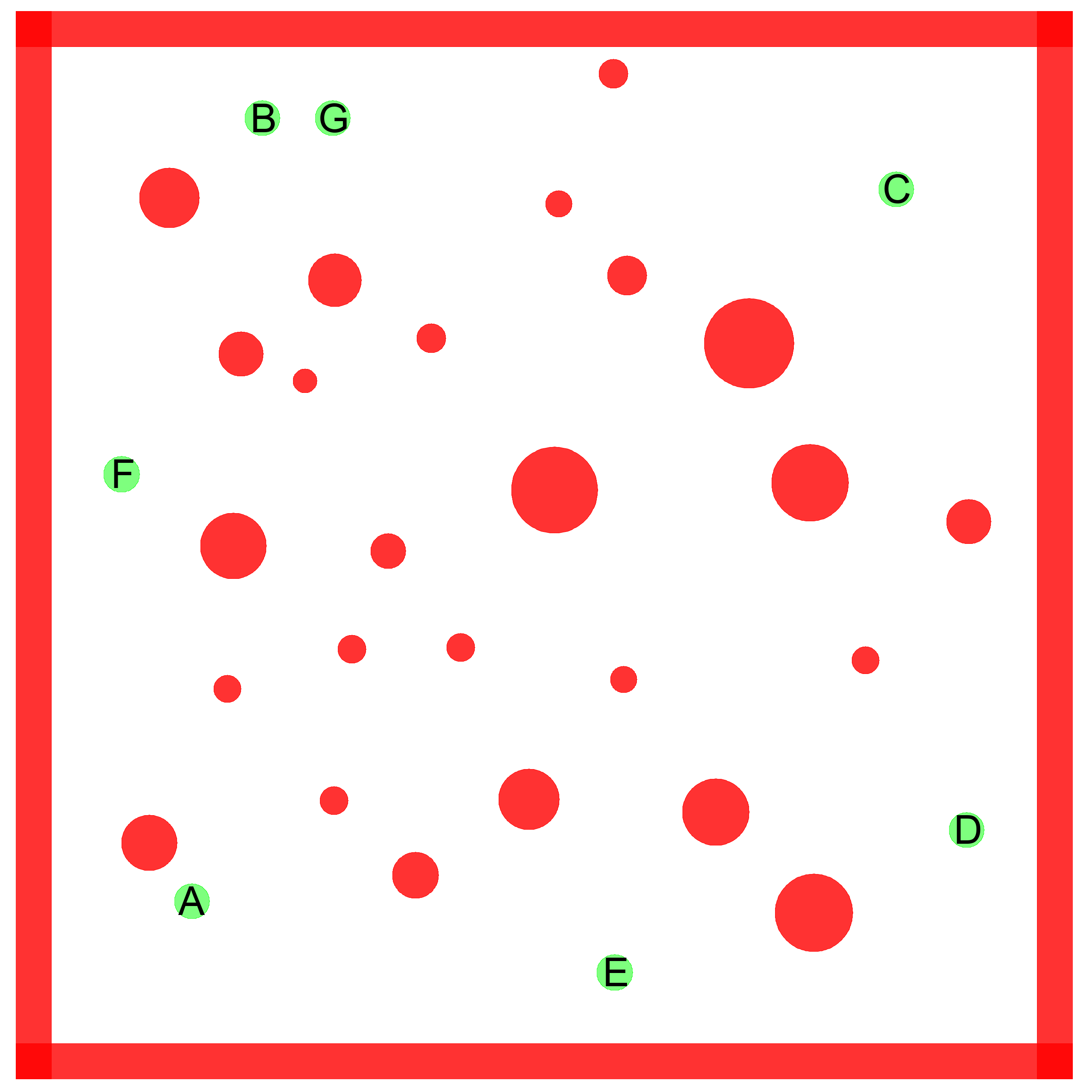}
  \caption{Forest}
  \label{fig:empty_envs_forest}
\end{subfigure}
\begin{subfigure}{.16\textwidth}
  \centering
  \includegraphics[width=.95\linewidth]{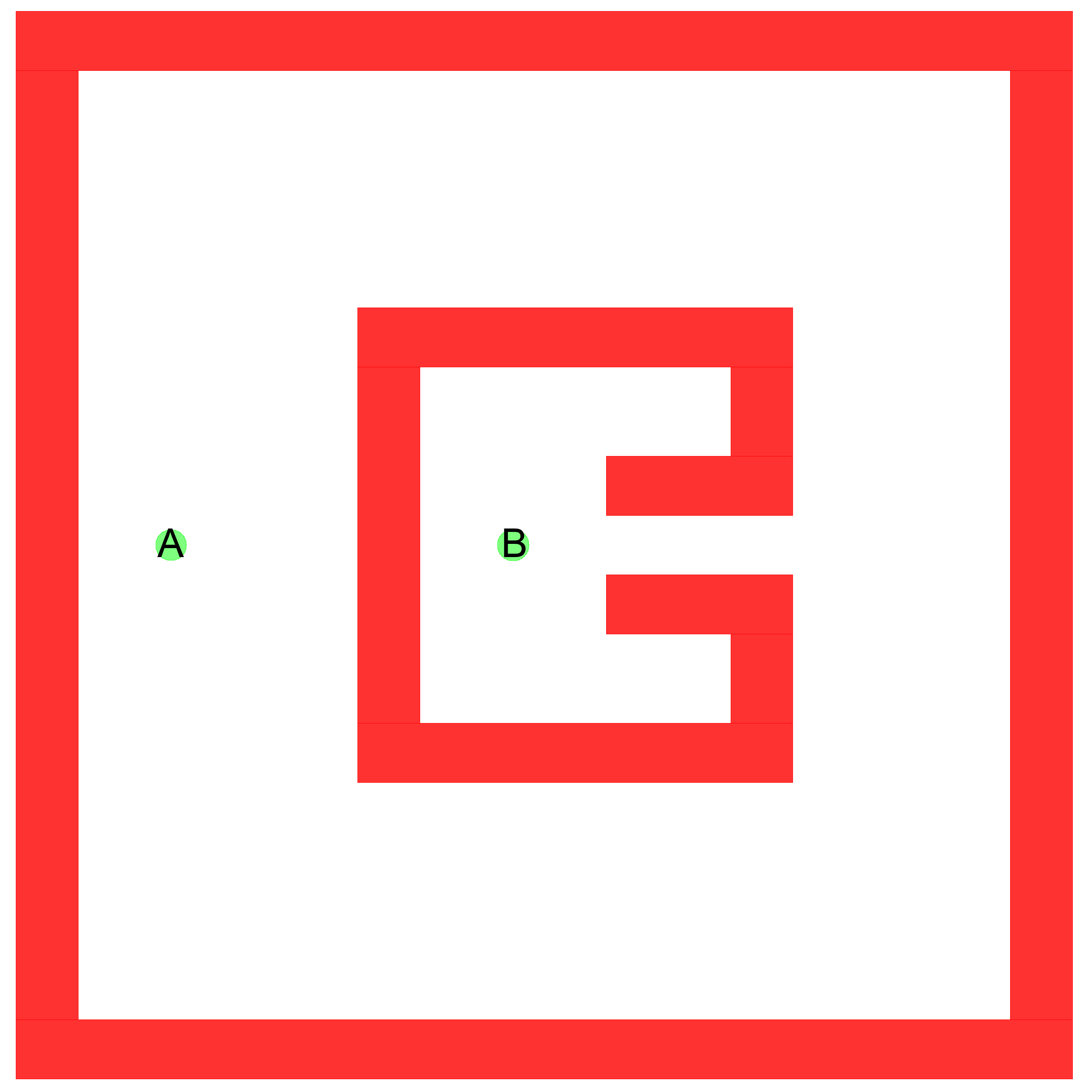}
  \caption{Bug Trap}
  \label{fig:empty_envs_bug_trap}
\end{subfigure}
\caption{\small Simulated environments used for experiments with the LTV-SDE and \mushr models. Letters indicate candidate starts and goals. The Simple Obstacle environment is a basic setup. Forest evaluates performance among many obstacles with some narrow passages. The Bug Trap is challenging due to the long, narrow passage.}
\vspace{-.2in}
\label{fig:empty_envs}
\end{figure}

\begin{figure*}[t]
\centering
\begin{subfigure}[b]{0.48\textwidth}
  \centering
  \includegraphics[width=\linewidth,height=3.5cm]{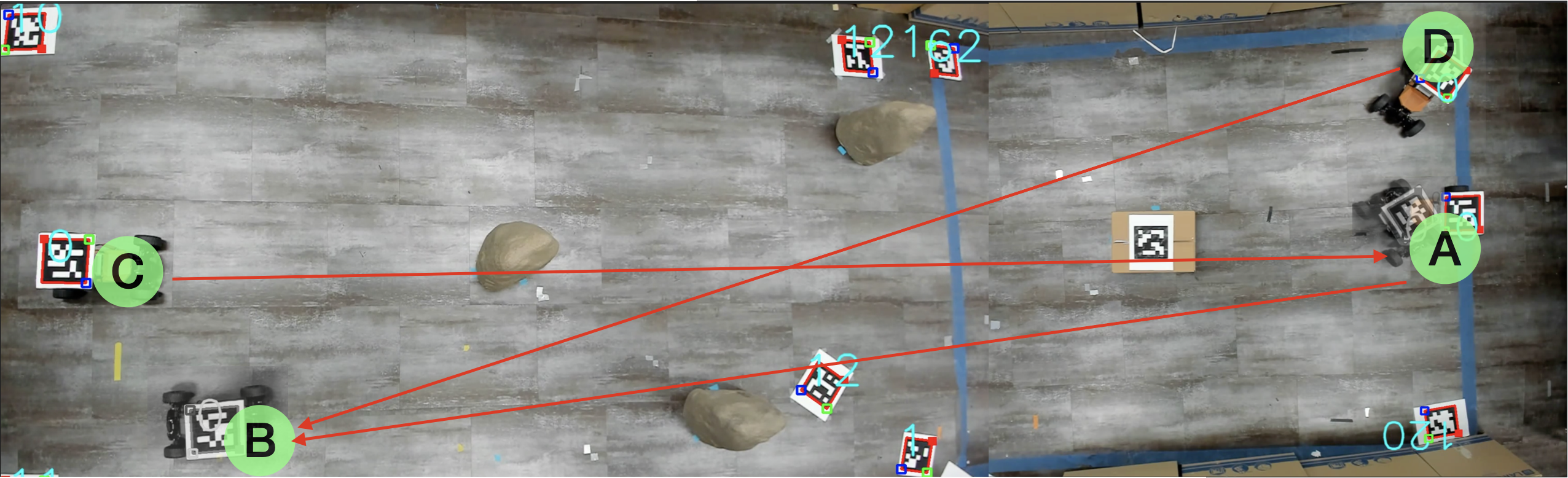}
  \label{fig:real_envs_multiple_obs1}
\end{subfigure}
\hfill
\begin{subfigure}[b]{0.48\textwidth}
  \centering
  \includegraphics[width=\linewidth,height=3.5cm]{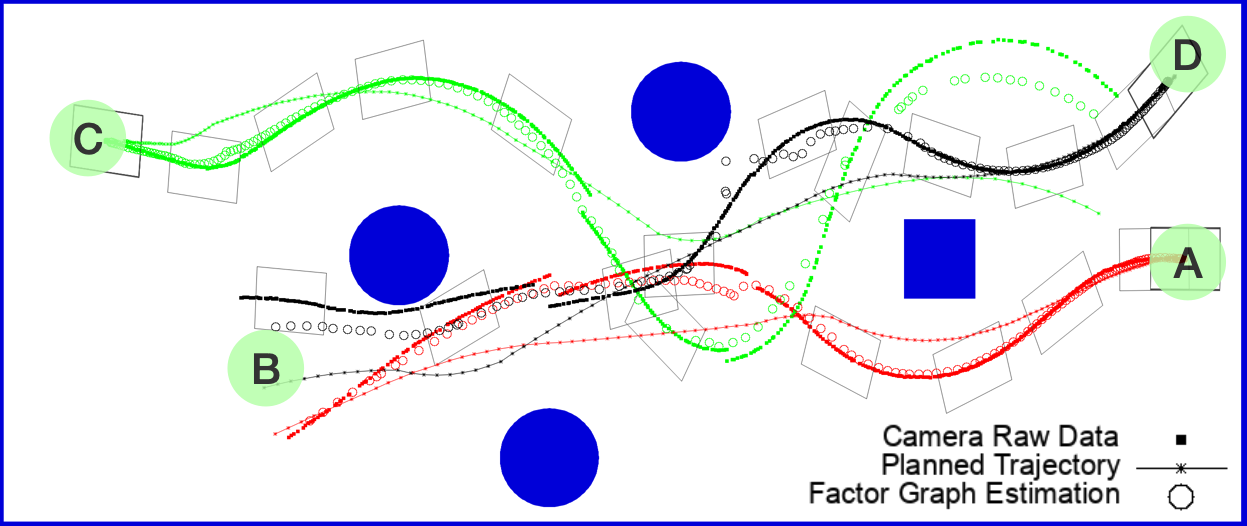}
  \label{fig:real_envs_multiple_obs2}
\end{subfigure}

\vspace{-.15in}
\caption*{Multiple obstacles}

\vspace{0.1cm}

\begin{subfigure}[b]{0.48\textwidth}
  \centering
  \includegraphics[width=\linewidth,height=3.5cm]{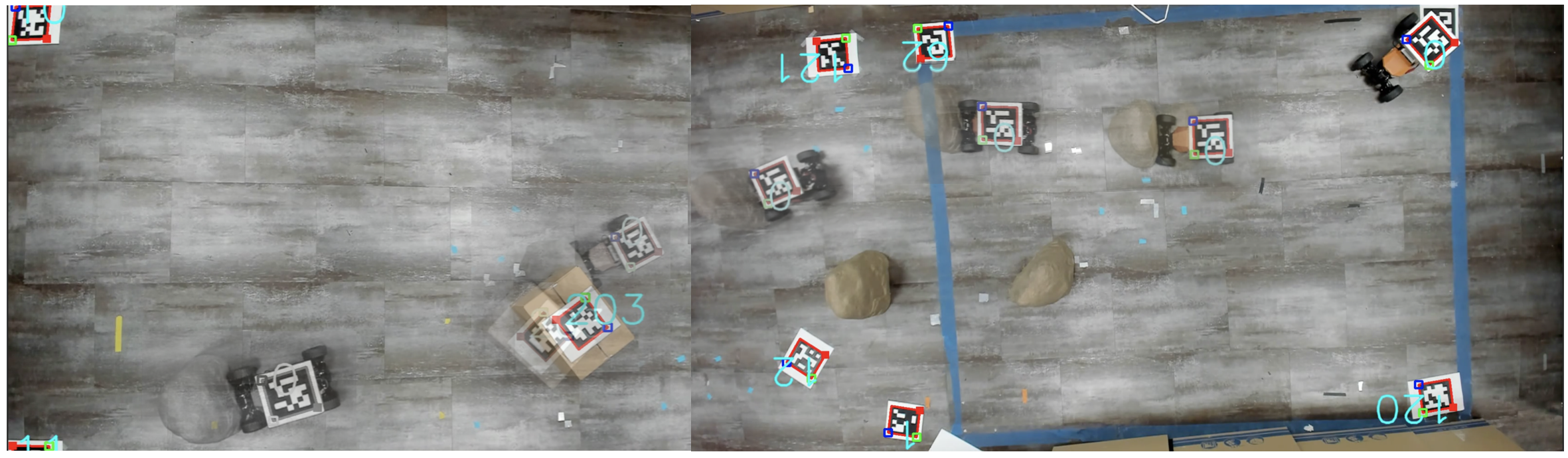}
  \label{fig:empty_envs_forest1}
\end{subfigure}
\hfill
\begin{subfigure}[b]{0.48\textwidth}
  \centering
  \includegraphics[width=\linewidth,height=3.5cm]{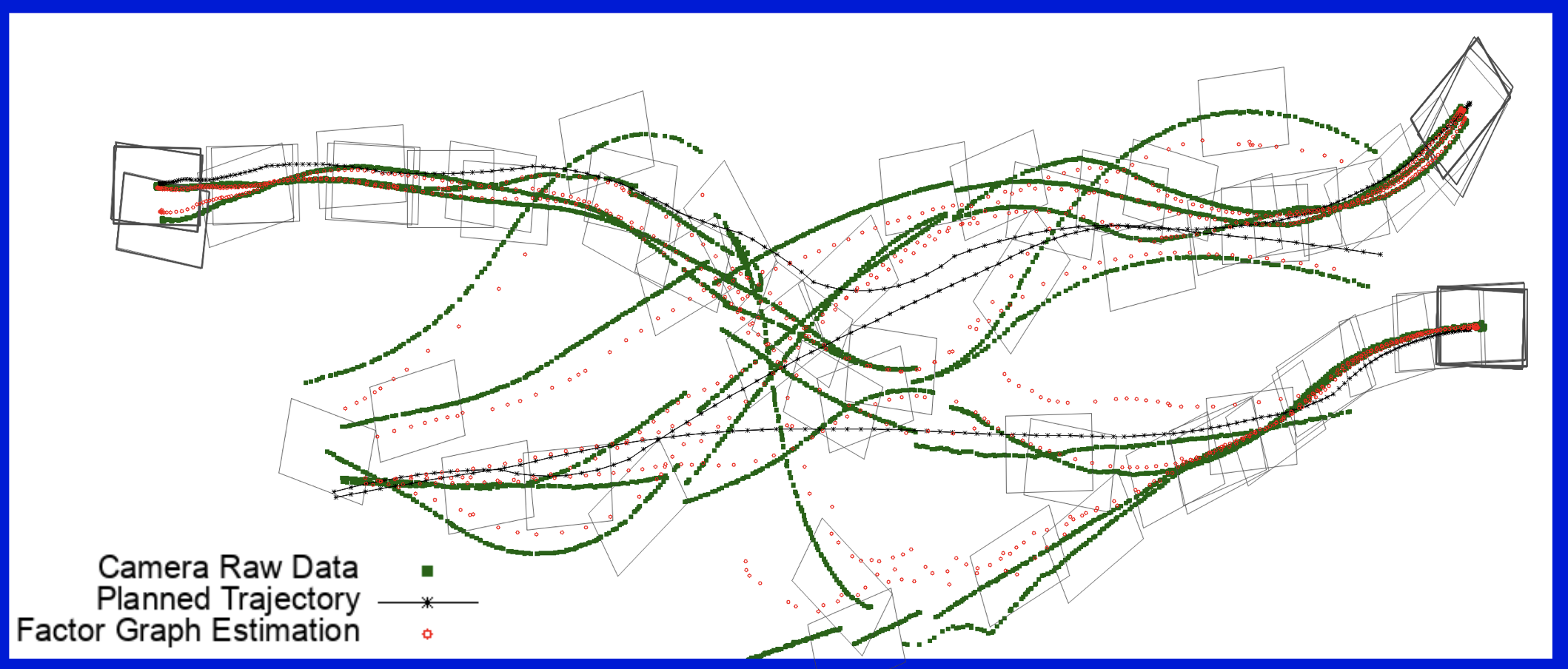}
  \label{fig:empty_envs_forest2}
\end{subfigure}
\vspace{-.15in}
\caption*{Movable boxes}
\vspace{-0.1in}

\caption{\small Experiments on a real \mushr. (Top) The robot navigates between (A-B), (C-A), and (D-B), avoiding obstacles. Initial poses in color, and final poses in gray. (Bottom) The robot follows a desired trajectory planned without obstacles. During execution, the environment has movable obstacles. Similar to the Ramp experiment of Figure \ref{fig:intro}, this setup tests \stela under partial observability.}
\vspace{-.1in}
\label{fig:empty_envs_real}
\end{figure*}

\section{EXPERIMENTS} \label{sec:results}

\subsection{Experimental setup}
\stela is tested in simulation in the environments shown in Fig.~\ref{fig:empty_envs} and with a real \mushr \cite{srinivasa2019mushr} robot in the environments shown in Fig.~\ref{fig:intro} and ~\ref{fig:empty_envs_real}. The system is tested against four levels of state space noise in simulation: $ \dot{x}_t=f(x_t,u_t)+N(0, I \cdot \sigma^x_i)$, and four levels of observation noise $z_t=h(x_t)+N(0, I \cdot \sigma^z_i)$, where $I$ is the identity matrix of appropriate dimension. Simulated systems are implemented as factor graphs using the GTSAM \cite{gtsam} library. Initial feasible trajectories are obtained using the AO-RRT kinodynamic planner \cite{kleinbort2020refined} from the ML4KP library. Both \stela and the comparison point \scate are implemented via GTSAM with Threading Building Blocks \cite{pheatt2008intel}, allowing parallelization for specific functions. Both algorithms are executed in a server with 72 cores, but each experiment is limited to 8 cores (the number of cores found in most computers) for fair comparison. 

All experiments are performed in a ROS-based system, where \stela and the comparing approaches are implemented as standalone nodes. All communication within nodes and the environment (either simulated or real) is performed through ROS messages, introducing additional unmodeled noise as time delays or lost messages. The update functions in \stela and \scate are implemented as ROS timers with a given frequency, which may not be respected due to extra computation time for the algorithm or from external sources.

\begin{figure}[]
    \centering
    \begin{subfigure}{.45\columnwidth}
        \vspace{-.15in}
        \centering
		\includegraphics[width=0.9\columnwidth]{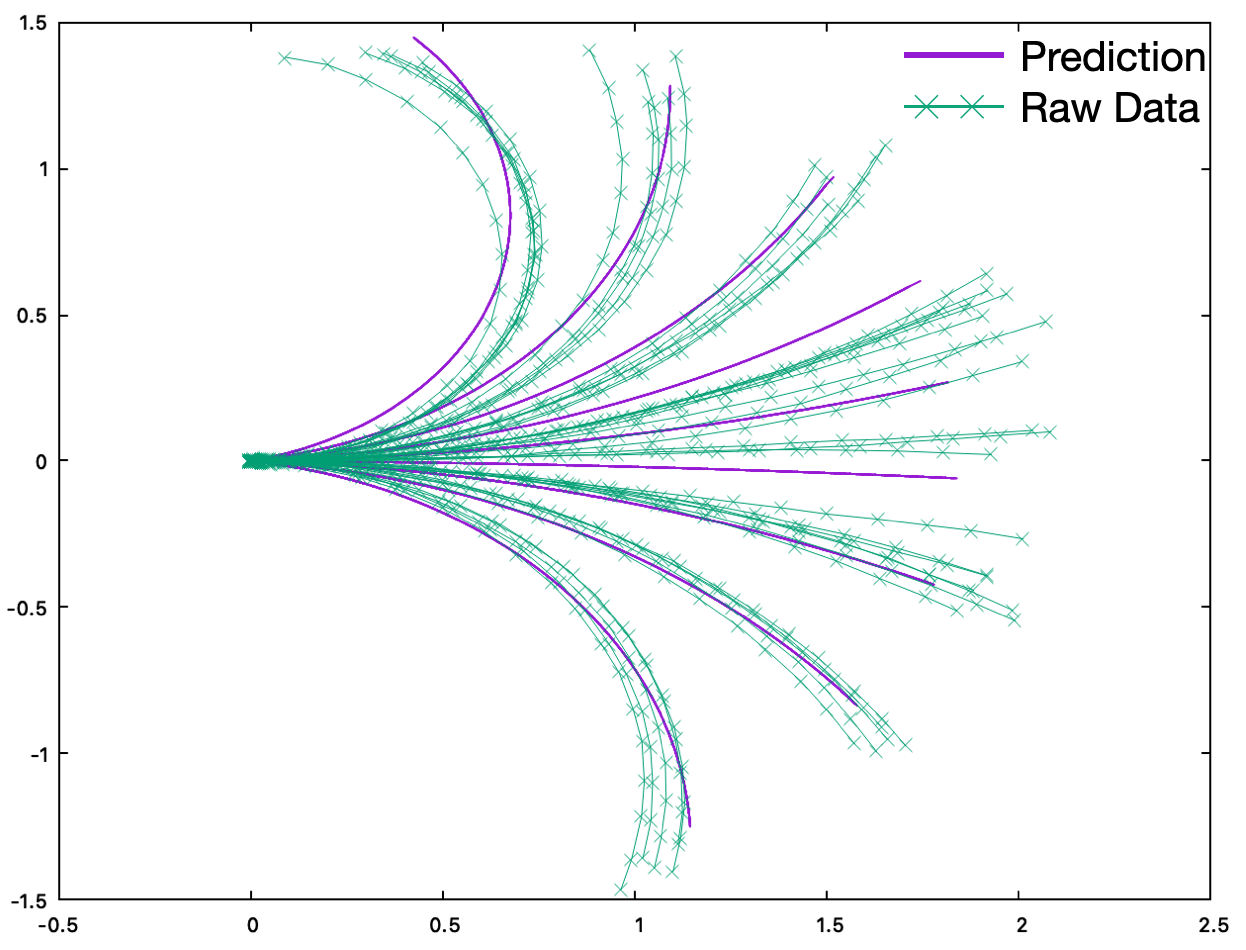}
        \caption{\mushr(real) SysId}
        \vspace{-.05in}
        \label{fig:sys_id_mushr}
    \end{subfigure}%
    \begin{subfigure}{.45\columnwidth}
        \centering
        \includegraphics[width=.9\columnwidth]{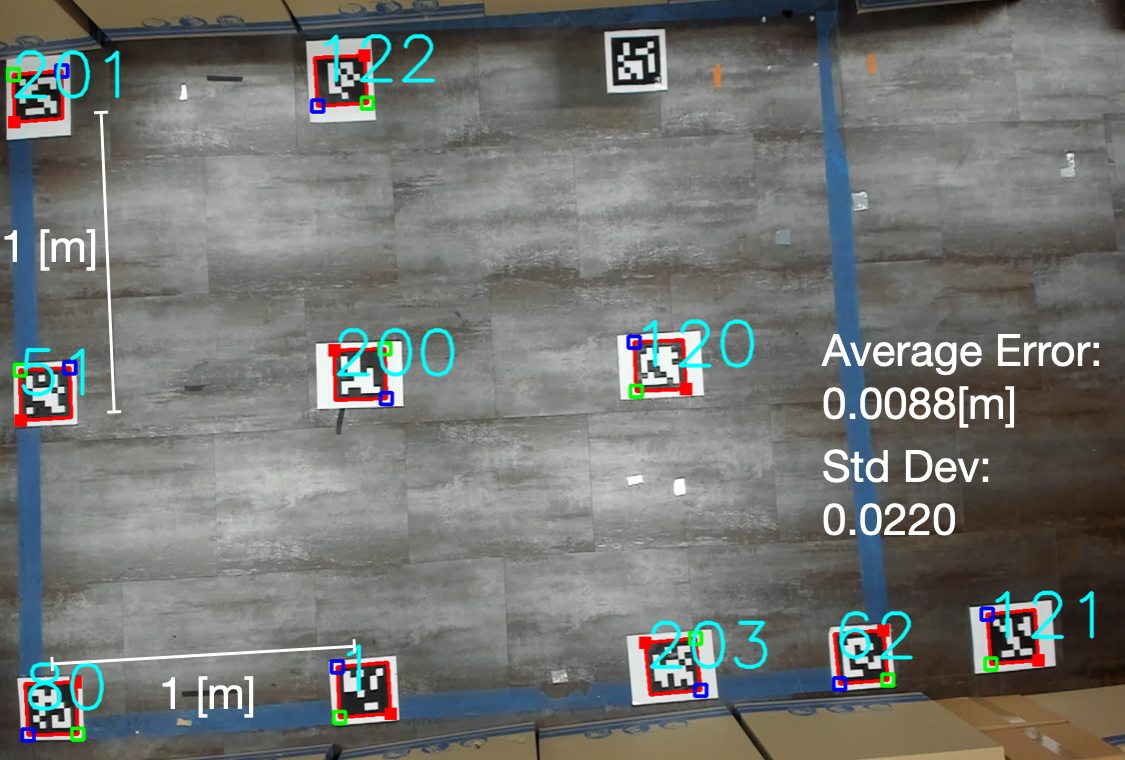}
        \caption{Observation Noise}
        \label{fig:real_z_noise}
    \end{subfigure}
    \caption{\small (Left) System Identification results for the real MuSHR. The model closely predicts the behavior of the robot along constant control trajectories, but a gap is still present. (Right) Observations $z_t$ are camera estimates of the robot's pose with the highest level of {\bf observation noise} ($\sigma^z_3$) chosen to match the real-world setup.}
    \vspace{-.2in}
    \label{fig:sys_id_plots}
\end{figure}

The {\bf robot systems} considered for experiments are:\\ 
\textbf{LTV-SDE} adopted from \cite{zheng2024cs} (eq. 50), $q,\dot{q} \in \mathbb{R}^2$. The controls $u\in [-0.2,0.2]\times[-0.2,0.2]$ represent acceleration.

\textbf{MuSHR} modeled as a second order system, where $q\in SE(2)$ and $\dot{q} \in \mathbb{R}^3$; the control $u\in \mathbb{R}^2$ corresponds to acceleration and steering angle. 

For the real MuSHR, a {\bf system identification} process to close the model gap is first performed. The parameters $\rho$ to be identified are the acceleration gain and the angular velocity gain.  In particular, using observed trajectories $\tau$ of the system under a known plan $p(T)$, the parameters $\rho$ are estimated so that the predictions given by the model best match the observations. An additional step was needed for the steering angle, as it has a non-linear bias. A polynomial fits the \textit{effective} steering angle vs the input.   
The system identification is solved via least squares optimization on a factor graph using the integration and dynamics factor. The controls correspond to the executed plan and are imposed via a prior factor. The steering angle in the trajectories is defined using a five-degree polynomial, which is fit from the same data. Figure~\ref{fig:sys_id_mushr} visualizes the raw data for the real platform and the prediction for constant control trajectories of fixed duration after system identification.  Asynchronous observations $\hat{x}_{i+\epsilon}$ of the robot's state are used for the sys. id process where $\epsilon \in [0,1]$ and assuming a noise distribution $\mathcal{N}(0,\sigma)$, between states $x_i$ and $x_{i+1}$. The initial guess is obtained by forward propagating $\hat{f}_{\rho_0}(x_0, u)$ for the duration of the plan.

The simulated MuSHR uses the same identified model as the real system with added noise in $\dot{q}$. Observations $z_t$ are camera estimates of the robot's pose. The highest level of {\bf observation noise} $\sigma^z_3$ matches the real-world setup, where ArUco markers~\cite{romero2018speeded} are positioned 1m apart on a grid (Fig.~\ref{fig:real_z_noise}). The mean error and the standard deviation of these pose estimation were measured given ground-truth.

\begin{table*}[]
\centering
\resizebox{\textwidth}{!}{%
\begin{tabular}{|c||cccccccccccccccc||}
\hline
              \multicolumn{17}{|c|}{LTV-SDE - Simple Obstacle}                                                                                                                                                                                                                                                                                                                                                                                                                                                                                                                                                                                      \\ \hline
             & \multicolumn{4}{c||}{Open Loop}                                                                                                                                                                & \multicolumn{4}{c||}{\scate-Naïve}                                                                                                              & \multicolumn{4}{c||}{\scate-SBMP}                                                                                                               & \multicolumn{4}{c||}{\stela}                                                                                              \\ \hline
             & \multicolumn{1}{c|}{$\sigma^x_0$}             & \multicolumn{1}{c|}{$\sigma^x_1$}             & \multicolumn{1}{c|}{$\sigma^x_2$}             & \multicolumn{1}{c||}{$\sigma^x_3$}             & \multicolumn{1}{c|}{$\sigma^x_0$} & \multicolumn{1}{c|}{$\sigma^x_1$} & \multicolumn{1}{c|}{$\sigma^x_2$} & \multicolumn{1}{c||}{$\sigma^x_3$} & \multicolumn{1}{c|}{$\sigma^x_0$} & \multicolumn{1}{c|}{$\sigma^x_1$} & \multicolumn{1}{c|}{$\sigma^x_2$} & \multicolumn{1}{c||}{$\sigma^x_3$} & \multicolumn{1}{c|}{$\sigma^x_0$} & \multicolumn{1}{c|}{$\sigma^x_1$} & \multicolumn{1}{c|}{$\sigma^x_2$} & $\sigma^x_3$  \\ \hline
$\sigma^z_0$ & \multicolumn{1}{c|}{1.0}                         & \multicolumn{1}{c|}{0.24}                         & \multicolumn{1}{c|}{0.0}                         & \multicolumn{1}{c||}{0.0}                         & \multicolumn{1}{c|}{1.0}             & \multicolumn{1}{c|}{1.0}             & \multicolumn{1}{c|}{1.0}             & \multicolumn{1}{c||}{0.75}             & \multicolumn{1}{c|}{0.88}             & \multicolumn{1}{c|}{0.88}             & \multicolumn{1}{c|}{0.88}             & \multicolumn{1}{c||}{0.88}             & \multicolumn{1}{c|}{1.0}          & \multicolumn{1}{c|}{1.0}          & \multicolumn{1}{c|}{1.0}          & 1.0           \\ \hline
$\sigma^z_1$ & \multicolumn{1}{c|}{\cellcolor[HTML]{343434}} & \multicolumn{1}{c|}{\cellcolor[HTML]{343434}} & \multicolumn{1}{c|}{\cellcolor[HTML]{343434}} & \multicolumn{1}{c||}{\cellcolor[HTML]{343434}} & \multicolumn{1}{c|}{1.0}             & \multicolumn{1}{c|}{1.0}             & \multicolumn{1}{c|}{0.75}             & \multicolumn{1}{c||}{0.50}             & \multicolumn{1}{c|}{1.0}             & \multicolumn{1}{c|}{1.0}             & \multicolumn{1}{c|}{1.0}             & \multicolumn{1}{c||}{0.88}             & \multicolumn{1}{c|}{1.0}          & \multicolumn{1}{c|}{1.0}          & \multicolumn{1}{c|}{1.0}          & 0.95          \\ \hline
$\sigma^z_2$ & \multicolumn{1}{c|}{\cellcolor[HTML]{343434}} & \multicolumn{1}{c|}{\cellcolor[HTML]{343434}} & \multicolumn{1}{c|}{\cellcolor[HTML]{343434}} & \multicolumn{1}{c||}{\cellcolor[HTML]{343434}} & \multicolumn{1}{c|}{1.0}             & \multicolumn{1}{c|}{1.0}             & \multicolumn{1}{c|}{0.75}             & \multicolumn{1}{c||}{0.75}             & \multicolumn{1}{c|}{1.0}             & \multicolumn{1}{c|}{1.0}             & \multicolumn{1}{c|}{0.88}             & \multicolumn{1}{c||}{0.75}             & \multicolumn{1}{c|}{1.0}          & \multicolumn{1}{c|}{1.0}          & \multicolumn{1}{c|}{1.0}          & 1.0           \\ \hline
$\sigma^z_3$ & \multicolumn{1}{c|}{\cellcolor[HTML]{343434}} & \multicolumn{1}{c|}{\cellcolor[HTML]{343434}} & \multicolumn{1}{c|}{\cellcolor[HTML]{343434}} & \multicolumn{1}{c||}{\cellcolor[HTML]{343434}} & \multicolumn{1}{c|}{0.75}             & \multicolumn{1}{c|}{1.0}             & \multicolumn{1}{c|}{1.0}             & \multicolumn{1}{c||}{1.0}             & \multicolumn{1}{c|}{1.0}             & \multicolumn{1}{c|}{1.0}             & \multicolumn{1}{c|}{1.0}             & \multicolumn{1}{c||}{0.88}             & \multicolumn{1}{c|}{1.0}          & \multicolumn{1}{c|}{1.0}          & \multicolumn{1}{c|}{0.95}         & 0.95          \\ \hline
\end{tabular}%
}
\caption{\small Success rates for the LVT-SDE on the Simple Obstacle scene for the different approaches. Columns correspond to different techniques and different levels of actuation noise. Rows correspond to different levels of observation noise.}
\label{tab:ltv_comp_so}
\end{table*}

\begin{table*}[]
\centering
\resizebox{\textwidth}{!}{\begin{tabular}{|c||cccccccccccccccccccc||}
\hline
              \multicolumn{21}{|c|}{LTV-SDE - Forest}                                                                                                                                                                                                                                                                                                                                                                                                                                                                                                                                                                                              \\ \hline
             & \multicolumn{4}{c||}{Open Loop}                                                                                                                                                                & \multicolumn{4}{c||}{SBMP Replanning}                                                                                                            & \multicolumn{4}{c||}{\scate-Naïve}                                                                                                                 & \multicolumn{4}{c||}{\scate-SBMP}                                                                                                                  & \multicolumn{4}{c||}{\stela}                                                                                             \\ \hline
             & \multicolumn{1}{c|}{$\sigma^x_0$}             & \multicolumn{1}{c|}{$\sigma^x_1$}             & \multicolumn{1}{c|}{$\sigma^x_2$}             & \multicolumn{1}{c||}{$\sigma^x_3$}             & \multicolumn{1}{c|}{$\sigma^x_0$} & \multicolumn{1}{c|}{$\sigma^x_1$}   & \multicolumn{1}{c|}{$\sigma^x_2$} & \multicolumn{1}{c||}{$\sigma^x_3$} & \multicolumn{1}{c|}{$\sigma^x_0$} & \multicolumn{1}{c|}{$\sigma^x_1$} & \multicolumn{1}{c|}{$\sigma^x_2$} & \multicolumn{1}{c||}{$\sigma^x_3$}     & \multicolumn{1}{c|}{$\sigma^x_0$} & \multicolumn{1}{c|}{$\sigma^x_1$} & \multicolumn{1}{c|}{$\sigma^x_2$} & \multicolumn{1}{c||}{$\sigma^x_3$}     & \multicolumn{1}{c|}{$\sigma^x_0$} & \multicolumn{1}{c|}{$\sigma^x_1$} & \multicolumn{1}{c|}{$\sigma^x_2$} & $\sigma^x_3$ \\ \hline
$\sigma^z_0$ & \multicolumn{1}{c|}{1.0}                      & \multicolumn{1}{c|}{0.0}                      & \multicolumn{1}{c|}{0.0}                      & \multicolumn{1}{c||}{0.0}                      & \multicolumn{1}{c|}{0.9}          & \multicolumn{1}{c|}{0.25}           & \multicolumn{1}{c|}{0.1}          & \multicolumn{1}{c||}{0.1}           & \multicolumn{1}{c|}{0.40}         & \multicolumn{1}{c|}{0.40}         & \multicolumn{1}{c|}{0.50}         & \multicolumn{1}{c||}{0.35}             & \multicolumn{1}{c|}{1.0}             & \multicolumn{1}{c|}{0.95}      & \multicolumn{1}{c|}{0.90}         & \multicolumn{1}{c||}{0.35}             & \multicolumn{1}{c|}{1.0}          & \multicolumn{1}{c|}{1.0}          & \multicolumn{1}{c|}{1.0}          & 0.96         \\ \hline
$\sigma^z_1$ & \multicolumn{1}{c|}{\cellcolor[HTML]{343434}} & \multicolumn{1}{c|}{\cellcolor[HTML]{343434}} & \multicolumn{1}{c|}{\cellcolor[HTML]{343434}} & \multicolumn{1}{c||}{\cellcolor[HTML]{343434}} & \multicolumn{1}{c|}{0.95}         & \multicolumn{1}{c|}{0.15}           & \multicolumn{1}{c|}{0.1}          & \multicolumn{1}{c||}{0.1}           & \multicolumn{1}{c|}{0.40}         & \multicolumn{1}{c|}{0.40}         & \multicolumn{1}{c|}{0.35}         & \multicolumn{1}{c||}{0.40}             & \multicolumn{1}{c|}{1.0}             & \multicolumn{1}{c|}{0.95}      & \multicolumn{1}{c|}{0.85}         & \multicolumn{1}{c||}{0.35}             & \multicolumn{1}{c|}{1.0}          & \multicolumn{1}{c|}{1.0}          & \multicolumn{1}{c|}{1.0}          & 0.96         \\ \hline
$\sigma^z_2$ & \multicolumn{1}{c|}{\cellcolor[HTML]{343434}} & \multicolumn{1}{c|}{\cellcolor[HTML]{343434}} & \multicolumn{1}{c|}{\cellcolor[HTML]{343434}} & \multicolumn{1}{c||}{\cellcolor[HTML]{343434}} & \multicolumn{1}{c|}{0.9}          & \multicolumn{1}{c|}{0.25}           & \multicolumn{1}{c|}{0.1}          & \multicolumn{1}{c||}{0.1}           & \multicolumn{1}{c|}{0.35}         & \multicolumn{1}{c|}{0.30}         & \multicolumn{1}{c|}{0.40}         & \multicolumn{1}{c||}{0.40}             & \multicolumn{1}{c|}{1.0}             & \multicolumn{1}{c|}{1.00}      & \multicolumn{1}{c|}{0.90}         & \multicolumn{1}{c||}{0.20}             & \multicolumn{1}{c|}{1.0}          & \multicolumn{1}{c|}{1.0}          & \multicolumn{1}{c|}{1.0}          & 0.98         \\ \hline
$\sigma^z_3$ & \multicolumn{1}{c|}{\cellcolor[HTML]{343434}} & \multicolumn{1}{c|}{\cellcolor[HTML]{343434}} & \multicolumn{1}{c|}{\cellcolor[HTML]{343434}} & \multicolumn{1}{c||}{\cellcolor[HTML]{343434}} & \multicolumn{1}{c|}{0.8}          & \multicolumn{1}{c|}{0.15}           & \multicolumn{1}{c|}{0.1}          & \multicolumn{1}{c||}{0.1}           & \multicolumn{1}{c|}{0.30}         & \multicolumn{1}{c|}{0.25}         & \multicolumn{1}{c|}{0.30}         & \multicolumn{1}{c||}{0.25}             & \multicolumn{1}{c|}{0.9}             & \multicolumn{1}{c|}{0.75}      & \multicolumn{1}{c|}{0.65}         & \multicolumn{1}{c||}{0.25}             & \multicolumn{1}{c|}{1.0}          & \multicolumn{1}{c|}{1.0}          & \multicolumn{1}{c|}{1.0}          & 0.92         \\ \hline
\end{tabular}%
}
\caption{\small Success rates for the LVT-SDE on Forest for the different approaches.}
\vspace{-0.1in}
\label{tab:ltv_comp_forest}
\end{table*}

\begin{table}[]
\centering
\resizebox{0.5\textwidth}{!}{\begin{tabular}{|c||cccccccccccc||}
\hline
  \multicolumn{13}{|c|}{LTV-SDE - Bug Trap}                                                                                      \\ \hline
 & \multicolumn{4}{c||}{Open Loop} & \multicolumn{4}{c||}{\scate-SBMP} & \multicolumn{4}{c||
 }{\stela} \\ \hline
 &
  \multicolumn{1}{c|}{$\sigma^x_0$} &
  \multicolumn{1}{c|}{$\sigma^x_1$} &
  \multicolumn{1}{c|}{$\sigma^x_2$} &
  \multicolumn{1}{c||}{$\sigma^x_3$} &
  \multicolumn{1}{c|}{$\sigma^x_0$} &
  \multicolumn{1}{c|}{$\sigma^x_1$} &
  \multicolumn{1}{c|}{$\sigma^x_2$} &
  \multicolumn{1}{c||}{$\sigma^x_3$} &
  \multicolumn{1}{c|}{$\sigma^x_0$} &
  \multicolumn{1}{c|}{$\sigma^x_1$} &
  \multicolumn{1}{c|}{$\sigma^x_2$} &
  $\sigma^x_3$ \\ \hline
$\sigma^z_0$ &
  \multicolumn{1}{c|}{1.0} &
  \multicolumn{1}{c|}{0.0} &
  \multicolumn{1}{c|}{0.0} &
  \multicolumn{1}{c||}{0.0} &
  \multicolumn{1}{c|}{1.0} &
  \multicolumn{1}{c|}{0.88} &
  \multicolumn{1}{c|}{0.88} &
  \multicolumn{1}{c||}{0} &
  \multicolumn{1}{c|}{1.0} &
  \multicolumn{1}{c|}{1.0} &
  \multicolumn{1}{c|}{1.0} &
  0.72 \\ \hline
$\sigma^z_1$ &
  \multicolumn{1}{c|}{\cellcolor[HTML]{343434}} &
  \multicolumn{1}{c|}{\cellcolor[HTML]{343434}} &
  \multicolumn{1}{c|}{\cellcolor[HTML]{343434}} &
  \multicolumn{1}{c||}{\cellcolor[HTML]{343434}} &
  \multicolumn{1}{c|}{1.0} &
  \multicolumn{1}{c|}{1.0} &
  \multicolumn{1}{c|}{0.88} &
  \multicolumn{1}{c||}{0.13} &
  \multicolumn{1}{c|}{1.0} &
  \multicolumn{1}{c|}{1.0} &
  \multicolumn{1}{c|}{1.0} &
  0.80 \\ \hline
$\sigma^z_2$ &
  \multicolumn{1}{c|}{\cellcolor[HTML]{343434}} &
  \multicolumn{1}{c|}{\cellcolor[HTML]{343434}} &
  \multicolumn{1}{c|}{\cellcolor[HTML]{343434}} &
  \multicolumn{1}{c||}{\cellcolor[HTML]{343434}} &
  \multicolumn{1}{c|}{1.0} &
  \multicolumn{1}{c|}{0.88} &
  \multicolumn{1}{c|}{0.62} &
  \multicolumn{1}{c||}{0} &
  \multicolumn{1}{c|}{1.0} &
  \multicolumn{1}{c|}{1.0} &
  \multicolumn{1}{c|}{0.96} &
  0.84 \\ \hline
$\sigma^z_3$ &
  \multicolumn{1}{c|}{\cellcolor[HTML]{343434}} &
  \multicolumn{1}{c|}{\cellcolor[HTML]{343434}} &
  \multicolumn{1}{c|}{\cellcolor[HTML]{343434}} &
  \multicolumn{1}{c||}{\cellcolor[HTML]{343434}} &
  \multicolumn{1}{c|}{1.0} &
  \multicolumn{1}{c|}{0.75} &
  \multicolumn{1}{c|}{0.75} &
  \multicolumn{1}{c||}{0.13} &
  \multicolumn{1}{c|}{1.0} &
  \multicolumn{1}{c|}{0.96} &
  \multicolumn{1}{c|}{0.92} &
  0.56 \\ \hline
\end{tabular}%
}
\caption{\small Success rates for the LVT-SDE on the Bug Trap. \scate-Naïve fails to initialize with a feasible plan and, therefore, is never successful. }
\vspace{-.1in}
\label{tab:ltv_comp_bug_trap}
\end{table}

\begin{table}[t]
\centering
\resizebox{0.45\textwidth}{!}{
\begin{tabular}{|c||cccc||cccc||}
\hline
             \multicolumn{9}{|c|}{\mushr (sim) - Simple Obstacle}  \\ \hline
             & \multicolumn{4}{c||}{Open Loop}                                                                                                                                                               & \multicolumn{4}{c||}{\stela}                                                                                               \\ \hline
             & \multicolumn{1}{c|}{$\sigma^x_0$}             & \multicolumn{1}{c|}{$\sigma^x_1$}             & \multicolumn{1}{c|}{$\sigma^x_2$}             & \multicolumn{1}{c||}{$\sigma^x_3$}             & \multicolumn{1}{c|}{$\sigma^x_0$} & \multicolumn{1}{c|}{$\sigma^x_1$} & \multicolumn{1}{c|}{$\sigma^x_2$} & $\sigma^x_3$ \\ \hline
$\sigma^z_0$ & \multicolumn{1}{c|}{1.0}                      & \multicolumn{1}{c|}{0.1}                         & \multicolumn{1}{c|}{0.0}                         & \multicolumn{1}{c||}{0.0}                         & \multicolumn{1}{c|}{1.0}          & \multicolumn{1}{c|}{1.0}          & \multicolumn{1}{c|}{0.80}         & 0.72         \\ \hline
$\sigma^z_1$ & \multicolumn{1}{c|}{\cellcolor[HTML]{343434}} & \multicolumn{1}{c|}{\cellcolor[HTML]{343434}} & \multicolumn{1}{c|}{\cellcolor[HTML]{343434}} & \multicolumn{1}{c||}{\cellcolor[HTML]{343434}} & \multicolumn{1}{c|}{1.0}          & \multicolumn{1}{c|}{1.0}          & \multicolumn{1}{c|}{0.80}         & 0.72         \\ \hline
$\sigma^z_2$ & \multicolumn{1}{c|}{\cellcolor[HTML]{343434}} & \multicolumn{1}{c|}{\cellcolor[HTML]{343434}} & \multicolumn{1}{c|}{\cellcolor[HTML]{343434}} & \multicolumn{1}{c||}{\cellcolor[HTML]{343434}} & \multicolumn{1}{c|}{1.0}          & \multicolumn{1}{c|}{1.0}          & \multicolumn{1}{c|}{0.80}         & 0.80         \\ \hline
$\sigma^z_3$ & \multicolumn{1}{c|}{\cellcolor[HTML]{343434}} & \multicolumn{1}{c|}{\cellcolor[HTML]{343434}} & \multicolumn{1}{c|}{\cellcolor[HTML]{343434}} & \multicolumn{1}{c||}{\cellcolor[HTML]{343434}} & \multicolumn{1}{c|}{0.96}         & \multicolumn{1}{c|}{1.0}          & \multicolumn{1}{c|}{0.80}         & 0.80         \\ \hline
\end{tabular}
}
\caption{\small Success rates for the sim. \mushr on Simple Obstacle.}
\label{tab:comparison_mushr_so}
\vspace{-.2in}
\end{table}

\noindent {\bf Comparison Points:} Table \ref{table:problems_reps} shows the number of problems (start-goal queries) per scene selected to test \stela against the alternatives. Multiple repetitions per query are performed. The baseline comparison point is open-loop execution of the desired trajectory. \scate was also chosen as a comparison point. \scate was originally implemented in MATLAB, and for consistency and performance purposes, it was re-implemented in C++, similar to \stela, reutilizing similar components. Two variations of \scate are considered. The first follows the original \scate by initializing the factor graph with a na\"ive straight-line plan from start to goal. The second variant is initialized with the same desired plan from the \sbmp as the proposed \stela approach. 

\begin{table}[h]
\vspace{-.15in}
\centering
\resizebox{0.5\columnwidth}{!}{%
\begin{tabular}{
>{\columncolor[HTML]{C0C0C0}}l |l|l|}
\cline{2-3}
\cellcolor[HTML]{FFFFFF}                                                             & \cellcolor[HTML]{C0C0C0}Trajs & \cellcolor[HTML]{C0C0C0}Reps \\ \hline
\multicolumn{1}{|l|}{\cellcolor[HTML]{C0C0C0}{\color[HTML]{000000} \small{Simple} Obst.}} & 1                                          & 5                                   \\ \hline
\multicolumn{1}{|l|}{\cellcolor[HTML]{C0C0C0}{\color[HTML]{000000} \small{Forest}}}          & 10                                         & 10                                  \\ \hline
\multicolumn{1}{|l|}{\cellcolor[HTML]{C0C0C0}{\color[HTML]{000000} \small{Bug Trap} }}        & 2                                          & 5                                   \\ \hline
\end{tabular}%
}
\caption{\small Number of trajectories and repetitions for each scene. }
\label{table:problems_reps}
\vspace{-.15in}
\end{table}

Since \scate requires a constant $dt$ between the states of the factor graph, the variable control durations of this tree had to be normalized by performing a weighted average operation over the edges and the states. Both implementations of \scate use the same obstacle factor as \stela and also include a factor to ensure that the controls are within the system's limits, which was observed to assist their performance. The \scate algorithm is only applicable to the LTV-SDE system but not to \mushr.

\begin{figure}
    \centering
    \includegraphics[width=\linewidth]{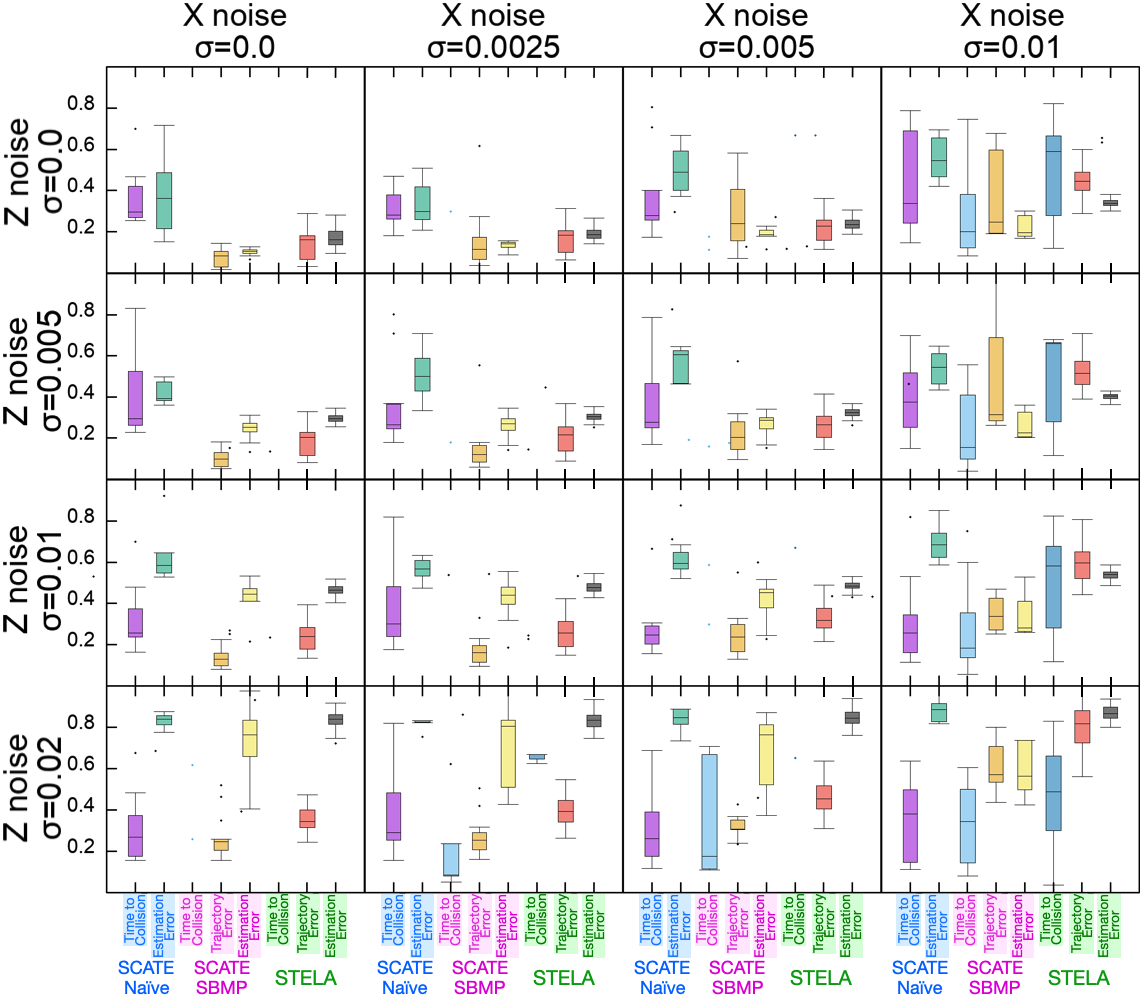}
    \caption{LTV-SDE Forest - Comparison of the Time to Collision, Normalized Trajectory Error, and Estimation Error between \scate-Naïve, \scate-SBMP, and \stela. Trajectory Error is skipped since \scate-Naïve does not have an initial trajectory to track.}
    \label{fig:ltv_alg_comp}
    \vspace{-.2in}
\end{figure}

\begin{figure}[h!]
\centering
\begin{subfigure}{.39\textwidth}
  \centering
  \includegraphics[width=\linewidth]{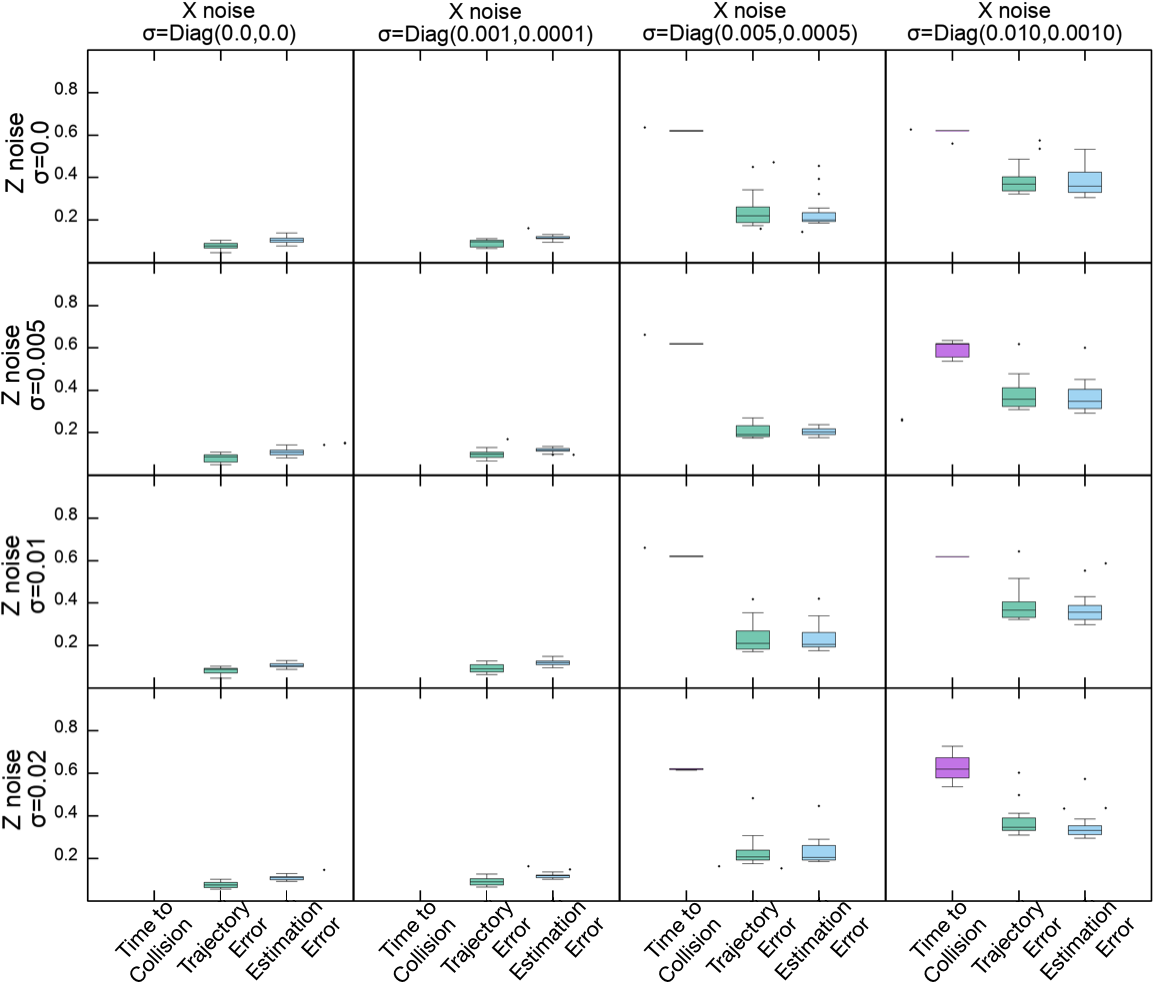}
  \caption{Simple Obstacle}
  \label{fig:plot_comp_mushr_simple_obstacle}
\end{subfigure}%
\\
\begin{subfigure}{.39\textwidth}
  \centering
  \includegraphics[width=\linewidth]{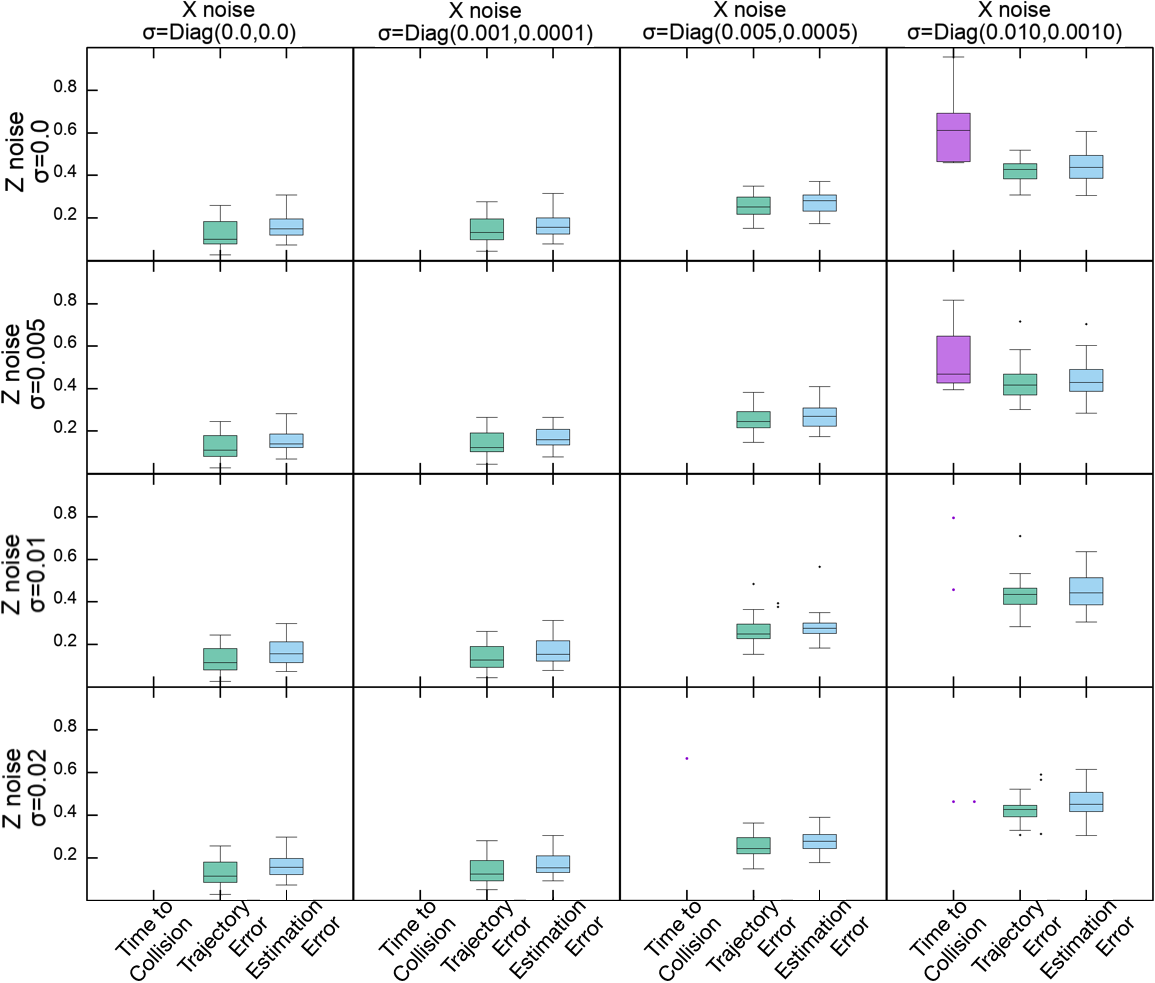}
  \caption{Forest}
  \label{fig:comparison_mushr_forest}
\end{subfigure}
\\
\begin{subfigure}{.39\textwidth}
  \centering
  \includegraphics[width=\linewidth]{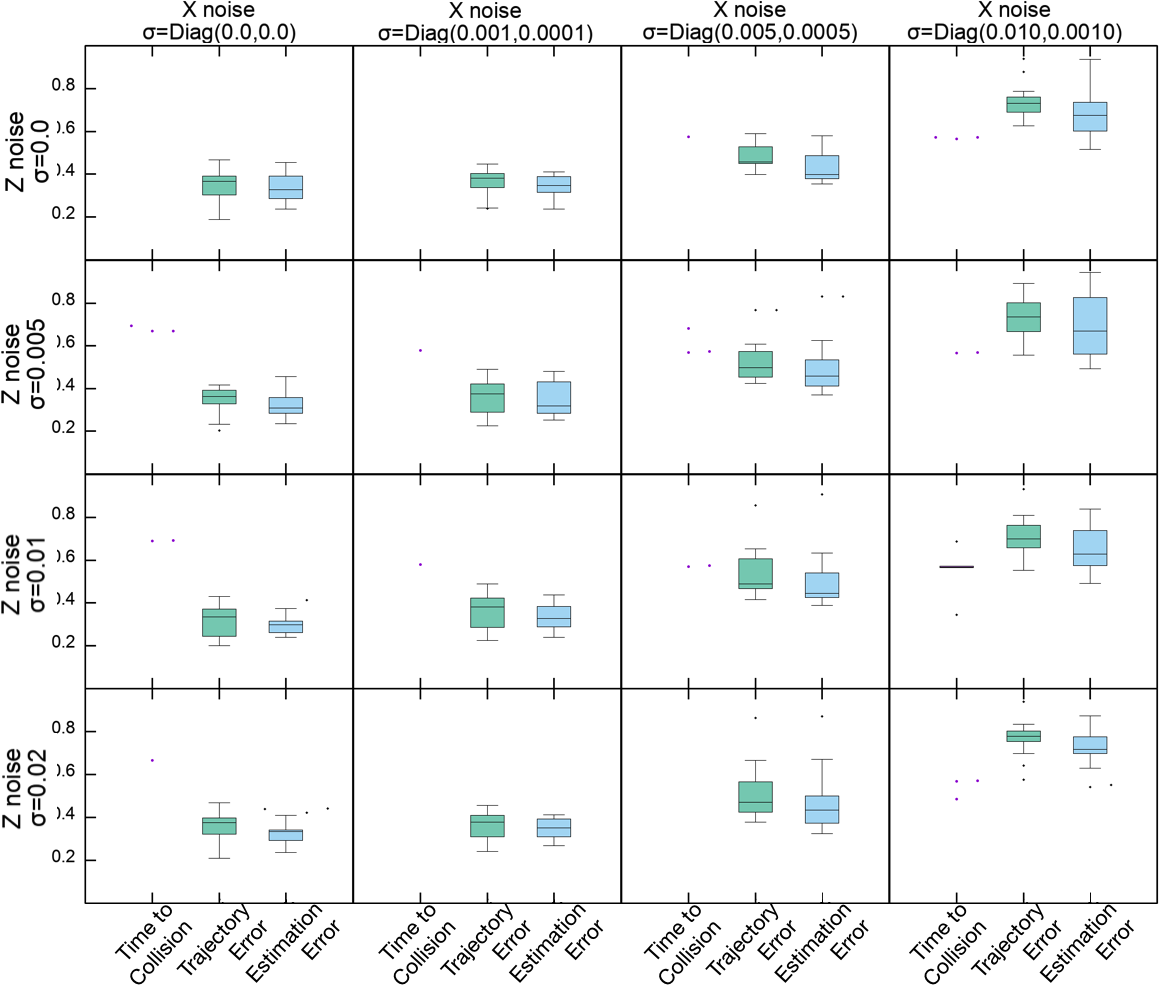}
  \caption{Bug Trap}
  \label{fig:comparison_mushr_bug_trap}
\end{subfigure}

\caption{\small \stela results for \mushr(sim). Three normalized metrics reported. \emph{Time to collision} is the rate of a trajectory traversed before a collision (no data if the success rate is 100\%). \emph{Trajectory error} (lower is better) is the L2-norm between the planned and executed trajectory, normalized over the highest error. \emph{Estimation error} (lower is better) is the L2-norm between the estimated trajectory and the ground truth, normalized over the highest error. \stela exhibits a very natural and slow degradation in performance as noise increases.}
\label{fig:empty_envs_results}
\end{figure}

\noindent {\bf Metrics:} Four metrics are considered during the evaluation.\\ \emph{Success: } Ratio of experiments where the robot reached the goal without collisions; the most critical metric for safety.\\ \emph{Cost: } Normalized duration of executed trajectories; normalized given the minimum duration solution per problem.\\ \emph{Estimation Error: } Error between ground truth and estimated trajectory; ground truth is not available for the real MuSHR.\\  \emph{Computation: } Time to perform an update for each algorithm; this metric reports only the time taken by the optimizer.

\begin{table}[t]
\centering
\resizebox{0.45\textwidth}{!}{
\begin{tabular}{|c||cccc||cccc||}
\hline
             \multicolumn{9}{|c|}{\mushr (sim) - Forest}                                                                                                                                                                                                                                                                                                                                                                                                                                                                                                                                                                                    \\ \hline
             & \multicolumn{4}{c||}{Open Loop}                                                                                                                                                                 & \multicolumn{4}{c||}{\stela}                                                                                               \\ \hline
             & \multicolumn{1}{c|}{$\sigma^x_0$}             & \multicolumn{1}{c|}{$\sigma^x_1$}             & \multicolumn{1}{c|}{$\sigma^x_2$}             & \multicolumn{1}{c||}{$\sigma^x_3$}             & \multicolumn{1}{c|}{$\sigma^x_0$} & \multicolumn{1}{c|}{$\sigma^x_1$} & \multicolumn{1}{c|}{$\sigma^x_2$} & $\sigma^x_3$ \\ \hline
$\sigma^z_0$ & \multicolumn{1}{c|}{1.0}                      & \multicolumn{1}{c|}{0.0}                      & \multicolumn{1}{c|}{0.0}                      & \multicolumn{1}{c||}{0.0}                      & \multicolumn{1}{c|}{1.0}          & \multicolumn{1}{c|}{1.0}          & \multicolumn{1}{c|}{1.00}         & 0.88         \\ \hline
$\sigma^z_1$ & \multicolumn{1}{c|}{\cellcolor[HTML]{343434}} & \multicolumn{1}{c|}{\cellcolor[HTML]{343434}} & \multicolumn{1}{c|}{\cellcolor[HTML]{343434}} & \multicolumn{1}{c||}{\cellcolor[HTML]{343434}} & \multicolumn{1}{c|}{1.0}          & \multicolumn{1}{c|}{1.0}          & \multicolumn{1}{c|}{0.96}         & 0.92         \\ \hline
$\sigma^z_2$ & \multicolumn{1}{c|}{\cellcolor[HTML]{343434}} & \multicolumn{1}{c|}{\cellcolor[HTML]{343434}} & \multicolumn{1}{c|}{\cellcolor[HTML]{343434}} & \multicolumn{1}{c||}{\cellcolor[HTML]{343434}} & \multicolumn{1}{c|}{1.0}          & \multicolumn{1}{c|}{1.0}          & \multicolumn{1}{c|}{0.98}         & 0.94         \\ \hline
$\sigma^z_3$ & \multicolumn{1}{c|}{\cellcolor[HTML]{343434}} & \multicolumn{1}{c|}{\cellcolor[HTML]{343434}} & \multicolumn{1}{c|}{\cellcolor[HTML]{343434}} & \multicolumn{1}{c||}{\cellcolor[HTML]{343434}} & \multicolumn{1}{c|}{0.98}         & \multicolumn{1}{c|}{1.0}          & \multicolumn{1}{c|}{1.00}         & 0.96         \\ \hline
\end{tabular}%
}
\caption{\small Success rates for the sim. \mushr on Forest.}
\label{tab:comparison_mushr_forest}
\vspace{-.1in}
\end{table}

\begin{table}[t]
\centering
\resizebox{0.45\textwidth}{!}{
\begin{tabular}{|c||cccc||cccc||}
\hline
             \multicolumn{9}{|c|}{\mushr (sim) - Bug Trap}                                                                                                                                                                                                                                                                                                                                                                                                                                                                                                                                                                                          \\ \hline
             & \multicolumn{4}{c||}{Open Loop}                                                                                                                                                                 & \multicolumn{4}{c||}{\stela}                                                                                               \\ \hline
             & \multicolumn{1}{c|}{$\sigma^x_0$}             & \multicolumn{1}{c|}{$\sigma^x_1$}             & \multicolumn{1}{c|}{$\sigma^x_2$}             & \multicolumn{1}{c||}{$\sigma^x_3$}             & \multicolumn{1}{c|}{$\sigma^x_0$} & \multicolumn{1}{c|}{$\sigma^x_1$} & \multicolumn{1}{c|}{$\sigma^x_2$} & $\sigma^x_3$ \\ \hline
$\sigma^z_0$ & \multicolumn{1}{c|}{1.0}                         & \multicolumn{1}{c|}{0.0}                         & \multicolumn{1}{c|}{0.0}                         & \multicolumn{1}{c||}{0.0}                         & \multicolumn{1}{c|}{1.00}         & \multicolumn{1}{c|}{1.00}         & \multicolumn{1}{c|}{0.95}         & 0.85         \\ \hline
$\sigma^z_1$ & \multicolumn{1}{c|}{\cellcolor[HTML]{343434}} & \multicolumn{1}{c|}{\cellcolor[HTML]{343434}} & \multicolumn{1}{c|}{\cellcolor[HTML]{343434}} & \multicolumn{1}{c||}{\cellcolor[HTML]{343434}} & \multicolumn{1}{c|}{0.85}         & \multicolumn{1}{c|}{0.90}         & \multicolumn{1}{c|}{0.80}         & 0.90         \\ \hline
$\sigma^z_2$ & \multicolumn{1}{c|}{\cellcolor[HTML]{343434}} & \multicolumn{1}{c|}{\cellcolor[HTML]{343434}} & \multicolumn{1}{c|}{\cellcolor[HTML]{343434}} & \multicolumn{1}{c||}{\cellcolor[HTML]{343434}} & \multicolumn{1}{c|}{0.90}         & \multicolumn{1}{c|}{0.95}         & \multicolumn{1}{c|}{0.80}         & 0.70         \\ \hline
$\sigma^z_3$ & \multicolumn{1}{c|}{\cellcolor[HTML]{343434}} & \multicolumn{1}{c|}{\cellcolor[HTML]{343434}} & \multicolumn{1}{c|}{\cellcolor[HTML]{343434}} & \multicolumn{1}{c||}{\cellcolor[HTML]{343434}} & \multicolumn{1}{c|}{0.95}         & \multicolumn{1}{c|}{1.00}         & \multicolumn{1}{c|}{0.90}         & 0.75         \\ \hline
\end{tabular}%

}
\caption{\small Success rates for the sim. \mushr on Bug Trap.}
\label{tab:comparison_mushr_bug_trap}
\vspace{-.1in}
\end{table}

\begin{figure}
\vspace{-.2in}
    \centering
    \includegraphics[width=\linewidth]{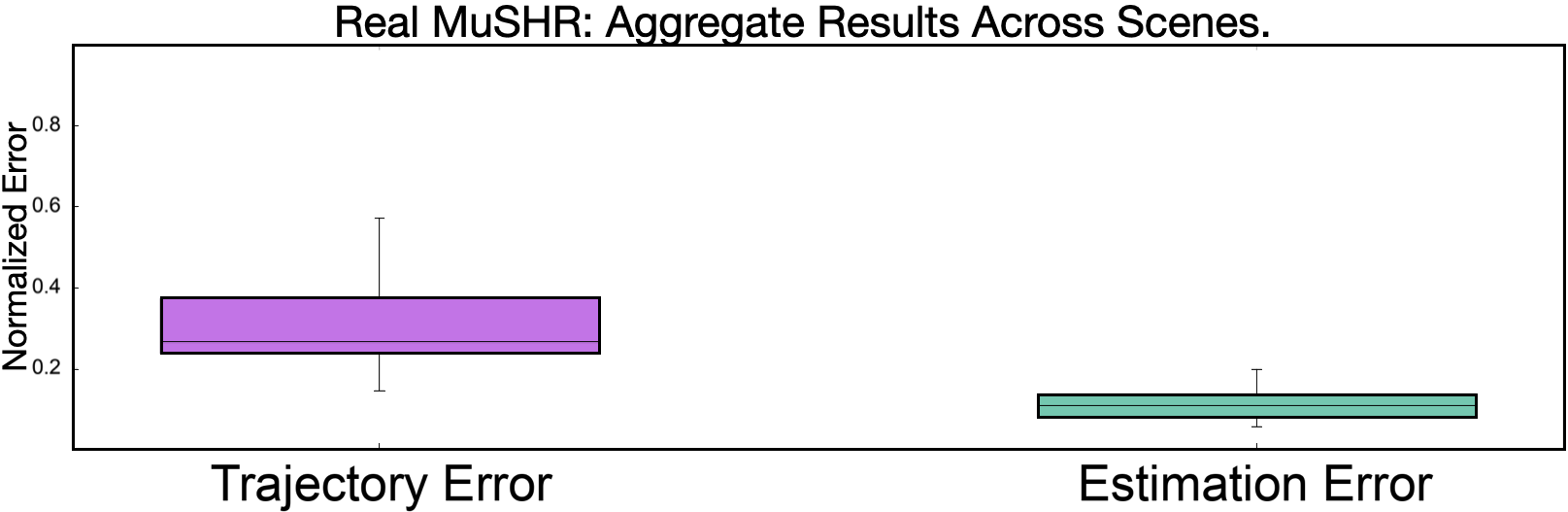}
    \caption{Aggregate results of the Normalized Trajectory Error and Estimation Error for \mushr(real) across scenes.}
    \label{fig:mushr_real_plot}
\end{figure}

\begin{figure}
    \centering
    \includegraphics[width=\linewidth]{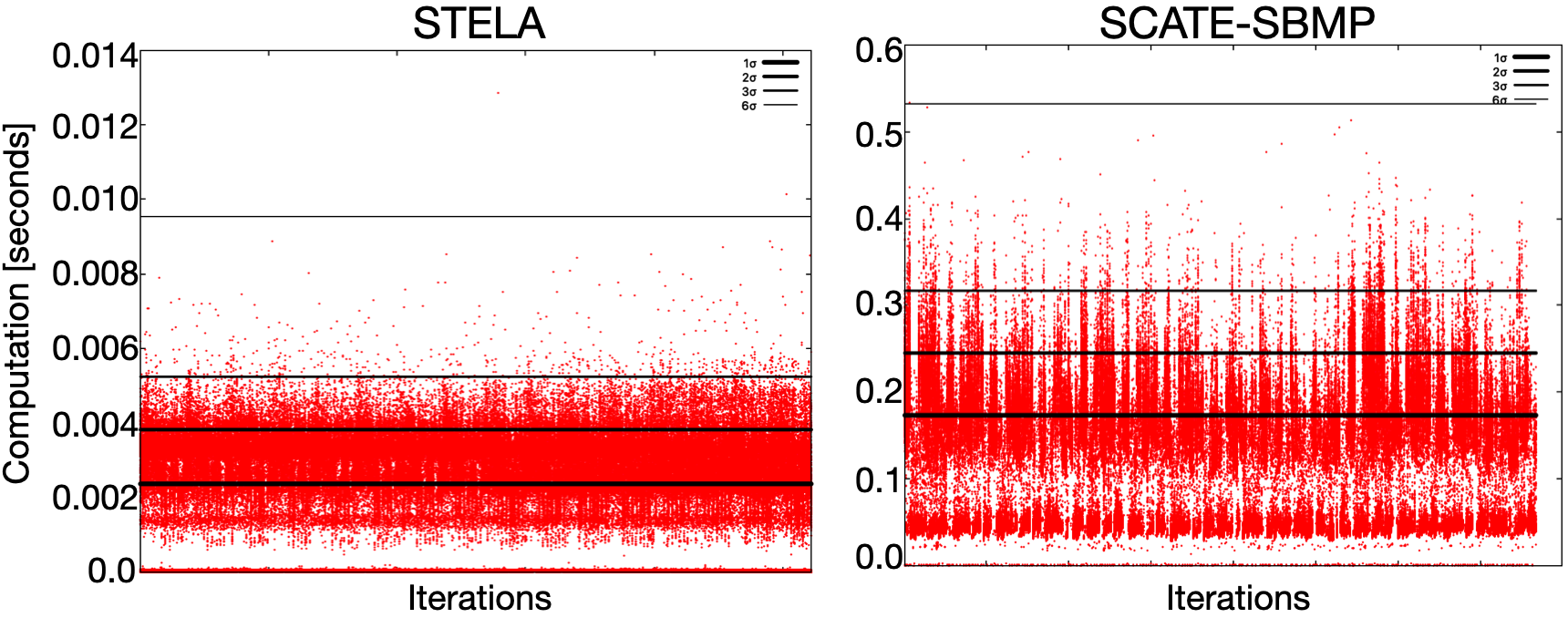}
    \caption{The computation time per call of \stela and \scate-SBMP shows a significant advantage in using the incremental computation and sliding window approach of \stela.}
    \label{fig:time_comp}
    \vspace{-.2in}
\end{figure}

\subsection{Simulation Results}
Tables \ref{tab:ltv_comp_so},\ref{tab:ltv_comp_forest}, and \ref{tab:ltv_comp_bug_trap} show the success rate of each algorithm per environment for the LTV-SDE system in simulation. Fig.~\ref{fig:ltv_alg_comp} presents the Time to Collision, Normalized Trajectory Error, and Estimation Error for the LTV-SDE Forest scenario.  Experiments with zero success rates are not included in the plots. Similarly, tables \ref{tab:comparison_mushr_so},\ref{tab:comparison_mushr_forest}, and \ref{tab:comparison_mushr_bug_trap} show success rates for the simulated \mushr (sim) alongside Figs. \ref{fig:empty_envs_results}. The computation time per iteration for the forest environment is shown in Fig.~\ref{fig:time_comp} for both \stela and \scate.

\openLoop showcases the effects of noise on the system's dynamics, resulting in collisions as soon as the minimum actuation noise level is introduced. No experiments were performed for non-zero observation noise since no estimation was performed in the case of the \openLoop baseline. 

\replanning is an online replanning strategy, which uses an informed, sampling-based, kinodynamic tree planner to generate new controls every second given the latest observation $z_t$ and the same dynamics model as \stela. This strategy is only tested in the forest environment for the LTV-SDE system (Table~\ref{tab:ltv_comp_forest}). This strategy degrades fast with higher dynamics noise.

\scate-Naïve works well in the Simple Obstacle environment for the LTV-SDE system. It faces significant challenges, however, in the Forest environment and fails to find any feasible paths in the Bug Trap environment, with no results reported. Initializing \scate with the feasible trajectory from the \sbmp allows for nearly 100\% success for the lowest actuation noise levels across all three environments for the LTV-SDE system. As actuation noise increases, however, the performance of \scate degrades. In setups with high actuation noise, both versions of \scate perform significantly worse than \stela.  As a reminder, the \scate comparison point is not directly applicable to the \mushr system.

\stela gets a 100\% success rate on the lowest noise levels while maintaining a high success rate on the most challenging levels across all environments. The increase in cost on the higher noise levels can be explained by the effect of the time variables in the factor graph optimization, which \textit{stretches} the edges as needed so as to avoid collisions and still solve problems. The \stela exhibits very similar behavior for the second-order, non-holonomic \mushr car as with the idealized, holonomic LTV-SDE system, which highlights its ability to work for different models of robotic systems.

\begin{table*}[t]
\centering
\resizebox{\textwidth}{!}{
\begin{tabular}{|c|c|ccc|c|cc|c|}
\hline
\textbf{}                       & \multirow{2}{*}{\stela} & \multicolumn{3}{c|}{\stela with variable window Sizes}                                                                                       & \multirow{2}{*}{Time not as Variable} & \multicolumn{2}{c|}{Obstacle Factor}                      & \multirow{2}{*}{Naive Initialization} \\ \cline{1-1} \cline{3-5} \cline{7-8}
\multicolumn{1}{|l|}{\textbf{}} &                         & \multicolumn{1}{l|}{10 Future - 0 Past Nodes} & \multicolumn{1}{l|}{1 Future - 10 Past Nodes} & \multicolumn{1}{l|}{1 Future - 0 Past Nodes} &                                       & \multicolumn{1}{l|}{None}      & \multicolumn{1}{l|}{PQP} &                                       \\ \hline
$(\sigma^x_1,\sigma^z_1)$       & 1.00                    & \multicolumn{1}{c|}{0.95}                     & \multicolumn{1}{c|}{0.78}                     & 0.80                                         & 0.95                                  & \multicolumn{1}{c|}{0.80}      & 1.0                      & 0.38                                  \\ \hline
$(\sigma^x_2,\sigma^z_2)$       & 1.00                    & \multicolumn{1}{c|}{0.95}                     & \multicolumn{1}{c|}{0.80}                     & 0.72                                         & 0.97                                  & \multicolumn{1}{c|}{0.90}      & 1.0                      & 0.37                                  \\ \hline
$(\sigma^x_3,\sigma^z_3)$       & 1.00                    & \multicolumn{1}{c|}{0.73}                     & \multicolumn{1}{c|}{0.45}                     & 0.41                                         & 0.77                                  & \multicolumn{1}{c|}{0.80}      & 0.98                     & 0.36                                  \\ \hline
$(\sigma^x_4,\sigma^z_4)$       & 0.96                    & \multicolumn{1}{c|}{0.42}                     & \multicolumn{1}{c|}{0.07}                     & 0.14                                         & 0.62                                  & \multicolumn{1}{c|}{0.64}      & 0.84                     &      0.33                        \\ \hline
\end{tabular}%

}
\caption{\small Ablation study of \stela. Comparisons are shown for (left to right): different sliding window sizes, no time factor variable, variations in the obstacle factors, and naive initialization (discretized straight line path from start to goal) instead of the \sbmp initialization.}
\label{tab:mushr_ablation}
\vspace{-.1in}
\end{table*}

\begin{figure*}
    \centering
    \begin{subfigure}{.38\textwidth}
      \centering
      \includegraphics[width=\columnwidth]{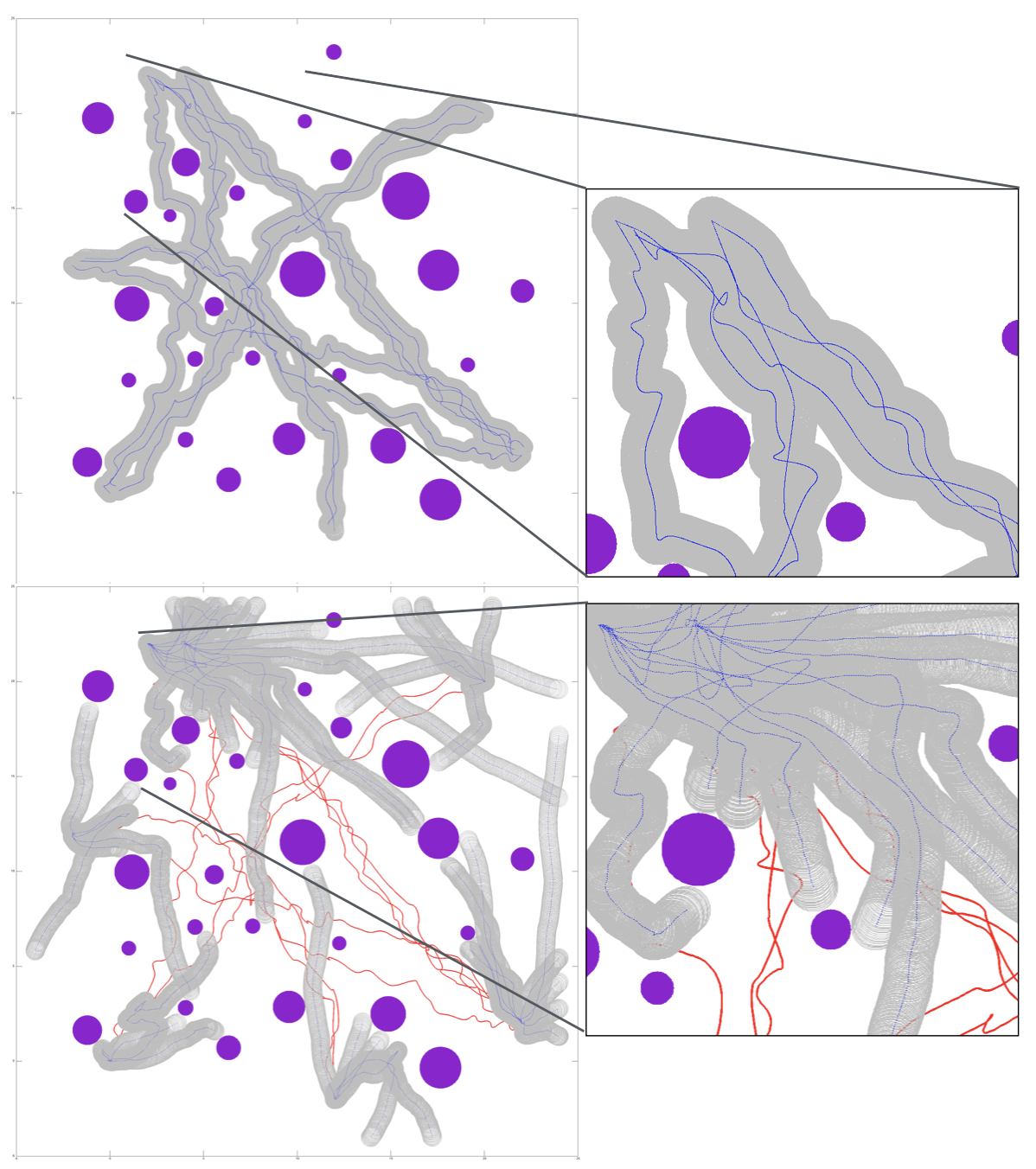}
      \caption{LTV: Open-loop}
        \label{fig:ltv_open_loop}
    \end{subfigure}
    \hfill
    \begin{subfigure}{.38\textwidth}
      \centering
      \includegraphics[width=\columnwidth]{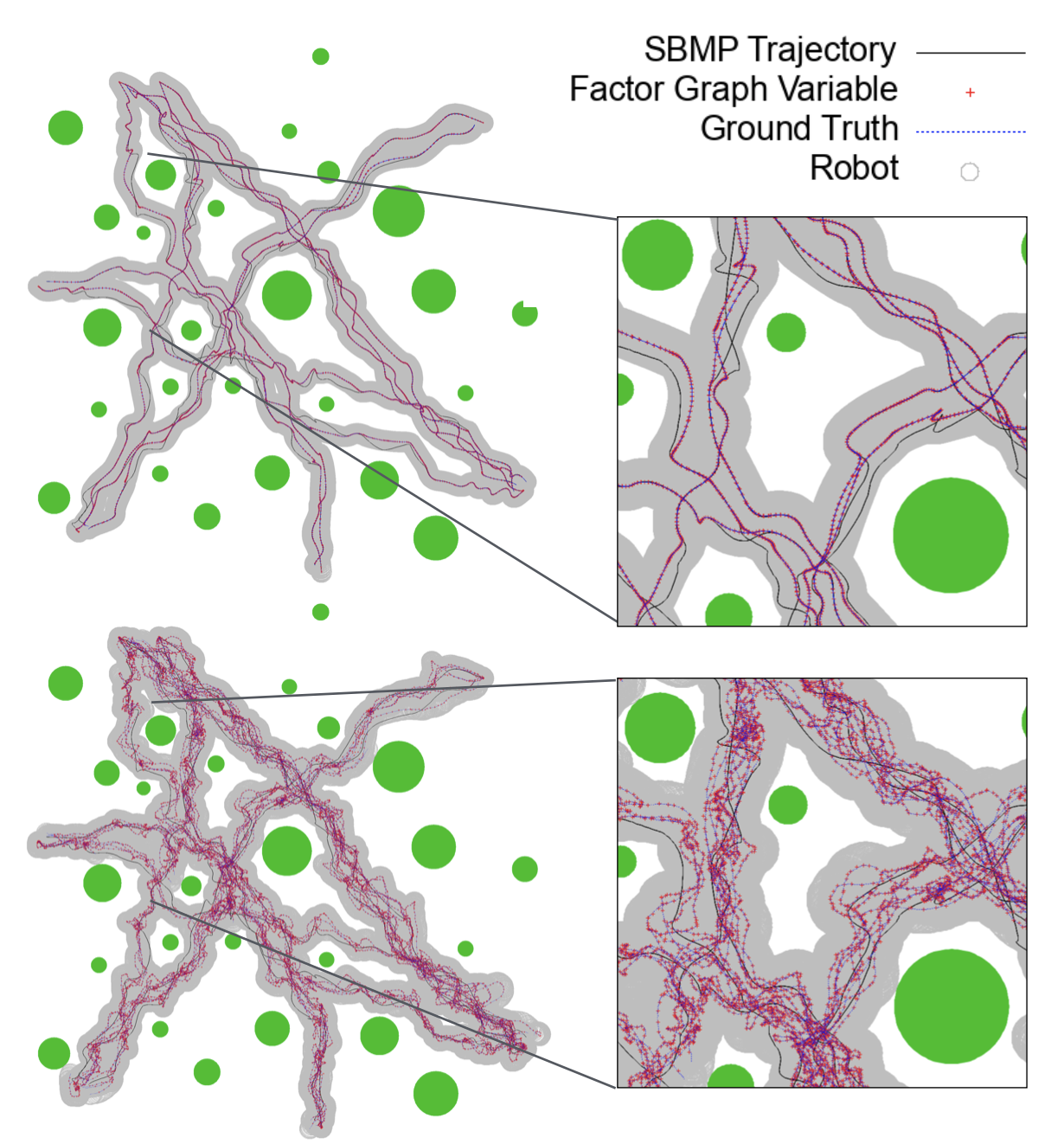}
      \caption{LTV: \stela}
    \label{fig:ltv_stela}
    \end{subfigure}
    \begin{subfigure}{.2\textwidth}
      \centering
      \includegraphics[height=8cm]{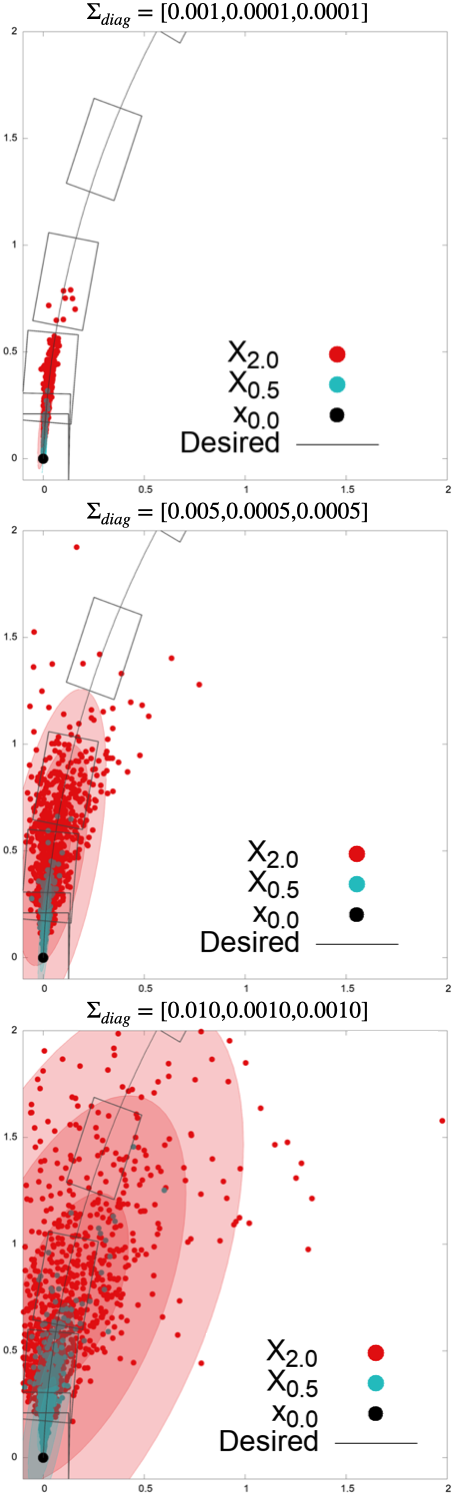}
      \caption{\mushr(sim): }
    \label{fig:mushr_open_loop}
    \end{subfigure}
    \caption{The effects of state-space noise in collision on the Forest environment for the Open-loop baseline (left) and the proposed \stela (middle). The top images correspond to the no-noise setup, while the bottom images correspond to the highest observation and dynamics noise ($\sigma^x_3$ and $\sigma^z_3$). Each image shows 100 trajectories (10 SBMP initializations with 10 repetitions each). \stela can still return collision-free solutions under significant noise, while open-loop execution of the desired trajectories results in collisions.
    (Right) For a desired trajectory (black line) of \mushr(Sim), confidence ellipses of $1\sigma$, $2\sigma$, and $3\sigma$ of forward propagating for 0.5s and 2s under the three levels (from top to bottom) of dynamics noise for 1K runs. }
    \vspace{-.2in}
\end{figure*}

\subsection{Real Experiments}
Real experiments are performed with a \mushr \cite{srinivasa2019mushr} robot in the scenes of Figures~\ref{fig:intro} and ~\ref{fig:empty_envs_real}.  The "multiple obstacles" environment is similar to the setups from simulated experiments, where collisions with obstacles are considered failures. The objective of the ``movable boxes" and ``ramp" environments is to test the ability to adapt to unmodeled environmental features. The second environment considers a set of \textit{movable} boxes that are not present during planning, and the robot can collide online without considering a failure. The third environment uses a ramp (not present during planning) that the robot needs to traverse to reach the goal. A total of 21 experiments were performed, nine on multiple obstacles (one collision), nine on movable obstacles, and three on the ramp environment (no collisions). Fig.~\ref{fig:empty_envs_real} shows qualitative results, while Fig.~\ref{fig:mushr_real_plot} reports the aggregated deviation from the planned trajectory (trajectory error) and the estimation error.

\subsection{\stela Ablation}
An ablation study is performed for \stela given the simulated \mushr system on the Forest environment. The ablation evaluation of the effect of the sliding window size, the use of the duration $\Delta T$ as a factor variable, the impact of the obstacle factor, as well as the impact of initializing with the \sbmp trajectory versus a naïve initialization. The size of the sliding window is divided into the future horizon and the past history. The reduction of either of these values from the default value of 10 in \stela results in performance degradation. This showcases the effects of performing trajectory estimation, i.e., smoothing, instead of filtering, as well as having a longer horizon than just computing the next control to be executed. \stela without time as a variable also performs worse, which shows the benefit of letting the optimizer adapt the duration of the edges so as to hit the desired states under the effects of noise.  \stela without the obstacle factor is more susceptible to collisions, even though it still uses prior factors that push the solution towards the desired, collision-free trajectory. The low-cost trajectories returned from the \sbmp are likely, however, to be in close proximity to obstacles, which makes following them susceptible to collisions without an obstacle factor. Two versions of the obstacle factor were tested: an SDF-based one, used by the default approach, and an obstacle factor for multiple obstacles within the distance threshold as computed online, given PQP distance calls. The multi-obstacle, PQP variant slightly underperforms the SDF one. Finally, a naïve initialization (straight-line to the goal) instead of the \sbmp trajectory initialization results in significant performance degradation. Frequently, no feasible solution is achieved given the initialization of the approach, and then the algorithm fails. The approach proceeds to work if a solution is found around the initialization.

\section{Discussion} \label{sec:discussion}

This paper presents \stela, a novel approach that seamlessly integrates the output of kinodynamic sampling-based motion planning with an integrated approach for trajectory estimation and following through factor graph optimization. \stela's effectiveness is highlighted by its ability to dynamically adapt plans based on real-time sensor data, resulting in improved accuracy in trajectory following and robustness against unmodeled environmental variations and noise. The experimental evaluations indicate that \stela  enhances the practical applicability of model-based planning and control methods. It also significantly outperforms alternatives in simulated evaluations as noise increases, while achieving desirable high-frequency control update rates. 

\section{Limitations} \label{sec:limitations}

\begin{wraptable}{r}{0.45\columnwidth}
\vspace{-.15in}
\centering
\resizebox{0.4\columnwidth}{!}{\begin{tabular}{|c||ccc||}
\hline
             & \multicolumn{3}{c|}{Mushr - Forest}    \\ \hline
             & \multicolumn{1}{c|}{$\sigma^x_4$}  & \multicolumn{1}{c|}{$\sigma^x_5$}   & \multicolumn{1}{c|}{$\sigma^x_6$}   \\ \hline
$\sigma^z_0$ & \multicolumn{1}{c|}{0.60}          & \multicolumn{1}{c|}{0.37}           & \multicolumn{1}{c|}{0.05}           \\ \hline
$\sigma^z_1$ & \multicolumn{1}{c|}{0.66}          & \multicolumn{1}{c|}{0.35}           & \multicolumn{1}{c|}{0.03}           \\ \hline
$\sigma^z_2$ & \multicolumn{1}{c|}{0.61}          & \multicolumn{1}{c|}{0.36}           & \multicolumn{1}{c|}{0.07}           \\ \hline
$\sigma^z_3$ & \multicolumn{1}{c|}{0.67}          & \multicolumn{1}{c|}{0.34}           & \multicolumn{1}{c|}{0.07}           \\ \hline
\end{tabular}%
}
\caption{\small \stela's success rates for the MuSHR(sim) robot for high dynamics noise.}
\vspace{-.10in}
\label{tab:comparison_mushr_extra}
\end{wraptable}

While \stela\ offers a significant increase in performance and computational efficiency, it shares some limitations of existing factor graph approaches. In the presence of extreme noise and deviation from the desired trajectory, \stela will fail to converge. To test the effects of significant noise, an additional experiment for the MuSHR(sim) robot evaluates \stela under higher levels of dynamics noise, as shown in Table~\ref{tab:comparison_mushr_extra}, where $\sigma^x_i \in [0.011,0.015,0.020]$ (along the columns). The extreme noise level $\sigma^x_6$ results mostly in failures, where 24\% of failures arise from \textit{Indeterminant Linear System Exception}, i.e., the accumulation of numerical errors, which does not occur for the lower noise levels; where the only failure mode corresponds to potential collisions. While there is degradation of performance as the noise levels increase, the earlier experiments have established that \stela outperforms alternatives and succeeds under reasonable noise levels as well as for the real-world setup.

Higher noise levels motivate the use of online replanning of the desired trajectory in parallel to executing \stela. Such replanning can be especially useful if there are solution paths along different homotopic classes. It may also be beneficial in the context of dynamic obstacles, which is a setup that was not tested in the current evaluation. \stela 's formulation, however, should naturally extend to such setups. 

\section{Acknowledgements} \label{sec:acknowledgements}

Work by the authors in this paper was partially supported by NSF NRT-FW-HTF award 2021628. Any opinions, findings and conclusions, or recommendations expressed in this material are those of the authors and do not necessarily reflect the views of the NSF. Kostas Bekris holds concurrent appointments as a Professor of Rutgers University and as an Amazon Scholar. This paper describes work performed at Rutgers University and is not associated with Amazon.

\bibliographystyle{styles/bibtex/splncs03}
\bibliography{refs.bib}

\begin{thebibliography}{10}
\providecommand{\url}[1]{\texttt{#1}}
\providecommand{\urlprefix}{URL }

\bibitem{agha2018slap}
Agha-mohammadi, A.a., Agarwal, S., Kim, S.K., Chakravorty, S., Amato, N.M.:
  Slap: Simultaneous localization and planning under uncertainty via dynamic
  replanning in belief space. IEEE Transactions on Robotics  34(5),  1195--1214
  (2018)

\bibitem{atreya2022state}
Atreya, P., Biswas, J.: State supervised steering function for sampling-based
  kinodynamic planning. arXiv preprint arXiv:2206.07227  (2022)

\bibitem{bahnemann2017sampling}
B{\"a}hnemann, R., Burri, M., Galceran, E., Siegwart, R., Nieto, J.:
  Sampling-based motion planning for active multirotor system identification.
  In: 2017 IEEE International Conference on Robotics and Automation (ICRA). pp.
  3931--3938. IEEE (2017)

\bibitem{barfoot2014batch}
Barfoot, T.D., Tong, C.H., S{\"a}rkk{\"a}, S.: Batch continuous-time trajectory
  estimation as exactly sparse gaussian process regression. In: Robotics:
  Science and Systems. vol.~10, pp. 1--10. Citeseer (2014)

\bibitem{cunningham2010ddf}
Cunningham, A., Paluri, M., Dellaert, F.: Ddf-sam: Fully distributed slam using
  constrained factor graphs. In: 2010 IEEE/RSJ International Conference on
  Intelligent Robots and Systems. pp. 3025--3030. IEEE (2010)

\bibitem{gtsam}
Dellaert, F., Contributors, G.: borglab/gtsam (May 2022),
  \url{https://github.com/borglab/gtsam)}

\bibitem{dellaert2017factor}
Dellaert, F., Kaess, M., et~al.: Factor graphs for robot perception.
  Foundations and Trends{\textregistered} in Robotics  6(1-2),  1--139 (2017)

\bibitem{forster2015imu}
Forster, C., Carlone, L., Dellaert, F., Scaramuzza, D.: Imu preintegration on
  manifold for efficient visual-inertial maximum-a-posteriori estimation. In:
  Robotics: Science and Systems XI (2015)

\bibitem{gammell2015batch}
Gammell, J.D., Srinivasa, S.S., Barfoot, T.D.: Batch informed trees (bit*):
  Sampling-based optimal planning via the heuristically guided search of
  implicit random geometric graphs. In: 2015 IEEE international conference on
  robotics and automation (ICRA). pp. 3067--3074. IEEE (2015)

\bibitem{goretkin2013optimal}
Goretkin, G., Perez, A., Platt, R., Konidaris, G.: Optimal sampling-based
  planning for linear-quadratic kinodynamic systems. In: 2013 IEEE
  International Conference on Robotics and Automation. pp. 2429--2436. IEEE
  (2013)

\bibitem{Hauser2020}
Hauser, K.: Motion and Path Planning, pp. 1--11. Springer Berlin Heidelberg,
  Berlin, Heidelberg (2020),
  \url{https://doi.org/10.1007/978-3-642-41610-1_165-1}

\bibitem{hoffmann2007autonomous}
Hoffmann, G.M., Tomlin, C.J., Montemerlo, M., Thrun, S.: Autonomous automobile
  trajectory tracking for off-road driving: Controller design, experimental
  validation and racing. In: 2007 American control conference. pp. 2296--2301.
  IEEE (2007)

\bibitem{honig2022db}
H{\"o}nig, W., Ortiz-Haro, J., Toussaint, M.: db-a*: Discontinuity-bounded
  search for kinodynamic mobile robot motion planning. In: 2022 IEEE/RSJ
  International Conference on Intelligent Robots and Systems (IROS). pp.
  13540--13547. IEEE (2022)

\bibitem{huang2017motion}
Huang, E., Mukadam, M., Liu, Z., Boots, B.: Motion planning with graph-based
  trajectories and gaussian process inference. In: 2017 IEEE International
  Conference on Robotics and Automation (ICRA). pp. 5591--5598. IEEE (2017)

\bibitem{jiang2021probabilistic}
Jiang, F., Marmon, A., De~Courten, I., Rasi, M., Dellaert, F.: Probabilistic
  tracking with deep factors. arXiv preprint arXiv:2112.01609  (2021)

\bibitem{johansson2000state}
Johansson, R., Robertsson, A., Nilsson, K., Verhaegen, M.: State-space system
  identification of robot manipulator dynamics. Mechatronics  10(3),  403--418
  (2000)

\bibitem{kaess2011bayes}
Kaess, M., Ila, V., Roberts, R., Dellaert, F.: The bayes tree: An algorithmic
  foundation for probabilistic robot mapping. In: Algorithmic Foundations of
  Robotics IX: Selected Contributions of the Ninth International Workshop on
  the Algorithmic Foundations of Robotics. pp. 157--173. Springer (2011)

\bibitem{kaess2012isam2}
Kaess, M., Johannsson, H., Roberts, R., Ila, V., Leonard, J.J., Dellaert, F.:
  isam2: Incremental smoothing and mapping using the bayes tree. The
  International Journal of Robotics Research  31(2),  216--235 (2012)

\bibitem{kaess2008isam}
Kaess, M., Ranganathan, A., Dellaert, F.: isam: Incremental smoothing and
  mapping. IEEE Transactions on Robotics  24(6),  1365--1378 (2008)

\bibitem{kalakrishnan2011stomp}
Kalakrishnan, M., Chitta, S., Theodorou, E., Pastor, P., Schaal, S.: Stomp:
  Stochastic trajectory optimization for motion planning. In: 2011 IEEE
  international conference on robotics and automation. pp. 4569--4574. IEEE
  (2011)

\bibitem{karaman2011sampling}
Karaman, S., Frazzoli, E.: Sampling-based algorithms for optimal motion
  planning. The international journal of robotics research  30(7),  846--894
  (2011)

\bibitem{king2022simultaneous}
King-Smith, M., Tsiotras, P., Dellaert, F.: Simultaneous control and trajectory
  estimation for collision avoidance of autonomous robotic spacecraft systems.
  In: 2022 International Conference on Robotics and Automation (ICRA). pp.
  257--264. IEEE (2022)

\bibitem{kleinbort2020refined}
Kleinbort, M., Granados, E., Solovey, K., Bonalli, R., Bekris, K.E., Halperin,
  D.: Refined analysis of asymptotically-optimal kinodynamic planning in the
  state-cost space. In: 2020 IEEE International Conference on Robotics and
  Automation (ICRA). pp. 6344--6350. IEEE (2020)

\bibitem{KschischangFG}
Kschischang, F., Frey, B., Loeliger, H.A.: Factor graphs and the sum-product
  algorithm. IEEE Transactions on Information Theory  47(2),  498--519 (2001)

\bibitem{larsen1999fast}
Larsen, E., Gottschalk, S., Lin, M.C., Manocha, D.: Fast proximity queries with
  swept sphere volumes. Tech. rep., Technical Report TR99-018, Department of
  Computer Science, University of North Carolina (1999)

\bibitem{lavalle2006planning}
LaValle, S.M.: Planning algorithms. Cambridge university press (2006)

\bibitem{littlefield2018efficient}
Littlefield, Z., Bekris, K.E.: Efficient and asymptotically optimal kinodynamic
  motion planning via dominance-informed regions. In: 2018 IEEE/RSJ
  International Conference on Intelligent Robots and Systems (IROS). pp. 1--9.
  IEEE (2018)

\bibitem{marcucci2023motion}
Marcucci, T., Petersen, M., von Wrangel, D., Tedrake, R.: Motion planning
  around obstacles with convex optimization. Science robotics  8(84),  eadf7843
  (2023)

\bibitem{mukadam2019steap}
Mukadam, M., Dong, J., Dellaert, F., Boots, B.: Steap: simultaneous trajectory
  estimation and planning. Autonomous Robots  43,  415--434 (2019)

\bibitem{mukadam2018continuous}
Mukadam, M., Dong, J., Yan, X., Dellaert, F., Boots, B.: Continuous-time
  gaussian process motion planning via probabilistic inference. The
  International Journal of Robotics Research  37(11),  1319--1340 (2018)

\bibitem{natarajan2021interleaving}
Natarajan, R., Choset, H., Likhachev, M.: Interleaving graph search and
  trajectory optimization for aggressive quadrotor flight. IEEE Robotics and
  Automation Letters  6(3),  5357--5364 (2021)

\bibitem{ortiz2024idb}
Ortiz-Haro, J., H{\"o}nig, W., Hartmann, V.N., Toussaint, M., Righetti, L.:
  idb-rrt: Sampling-based kinodynamic motion planning with motion primitives
  and trajectory optimization. arXiv preprint arXiv:2403.10745  (2024)

\bibitem{paden2016survey}
Paden, B., {\v{C}}{\'a}p, M., Yong, S.Z., Yershov, D., Frazzoli, E.: A survey
  of motion planning and control techniques for self-driving urban vehicles.
  IEEE Transactions on intelligent vehicles  1(1),  33--55 (2016)

\bibitem{paden2020generalized}
Paden, B., Frazzoli, E.: A generalized label correcting method for optimal
  kinodynamic motion planning. In: Algorithmic Foundations of Robotics XII:
  Proceedings of the Twelfth Workshop on the Algorithmic Foundations of
  Robotics. pp. 512--527. Springer (2020)

\bibitem{pheatt2008intel}
Pheatt, C.: Intel{\textregistered} threading building blocks. Journal of
  Computing Sciences in Colleges  23(4),  298--298 (2008)

\bibitem{ratliff2009chomp}
Ratliff, N., Zucker, M., Bagnell, J.A., Srinivasa, S.: Chomp: Gradient
  optimization techniques for efficient motion planning. In: 2009 IEEE
  international conference on robotics and automation. pp. 489--494. IEEE
  (2009)

\bibitem{romero2018speeded}
Romero-Ramirez, F.J., Mu{\~n}oz-Salinas, R., Medina-Carnicer, R.: Speeded up
  detection of squared fiducial markers. Image and vision Computing  76,
  38--47 (2018)

\bibitem{schulman2013finding}
Schulman, J., Ho, J., Lee, A.X., Awwal, I., Bradlow, H., Abbeel, P.: Finding
  locally optimal, collision-free trajectories with sequential convex
  optimization. In: RSS. vol.~9, pp. 1--10. Berlin, Germany (2013)

\bibitem{sola2018micro}
Sola, J., Deray, J., Atchuthan, D.: A micro lie theory for state estimation in
  robotics. arXiv preprint arXiv:1812.01537  (2018)

\bibitem{srinivasa2019mushr}
Srinivasa, S.S., Lancaster, P., Michalove, J., Schmittle, M., Summers, C.,
  Rockett, M., Scalise, R., Smith, J.R., Choudhury, S., Mavrogiannis, C.,
  et~al.: Mushr: A low-cost, open-source robotic racecar for education and
  research. arXiv preprint arXiv:1908.08031  (2019)

\bibitem{toussaint2009robot}
Toussaint, M.: Robot trajectory optimization using approximate inference. In:
  Proceedings of the 26th annual international conference on machine learning.
  pp. 1049--1056 (2009)

\bibitem{toussaint2014newton}
Toussaint, M.: Newton methods for k-order markov constrained motion problems.
  arXiv:1407.0414  (2014)

\bibitem{webb2013kinodynamic}
Webb, D.J., Van Den~Berg, J.: Kinodynamic rrt*: Asymptotically optimal motion
  planning for robots with linear dynamics. In: 2013 IEEE international
  conference on robotics and automation. pp. 5054--5061. IEEE (2013)

\bibitem{sysIdGPTingfan}
Wu, T., Movellan, J.: Semi-parametric gaussian process for robot system
  identification. In: 2012 IEEE/RSJ International Conference on Intelligent
  Robots and Systems. pp. 725--731 (2012)

\bibitem{zheng2024cs}
Zheng, D., Ridderhof, J., Zhang, Z., Tsiotras, P., Agha-mohammadi, A.A.:
  Cs-brm: A probabilistic roadmap for consistent belief space planning with
  reachability guarantees. IEEE Transactions on Robotics  (2024)

\end{thebibliography}
\end{document}